\documentclass[10pt]{article} 

\usepackage[accepted]{rlj} 

\usepackage{amssymb}            
\usepackage{mathrsfs}           
\usepackage{graphicx}           
\usepackage{subcaption}         
\usepackage[space]{grffile}     
\usepackage{url}                
\usepackage{lipsum}             
\usepackage{booktabs}
\usepackage{rotating}
\usepackage{tabularray}
\UseTblrLibrary{diagbox}
\usepackage{multirow}
\usepackage{makecell}
\usepackage{xcolor}
\usepackage{amsmath,bm}

\definecolor{ultramarine}{RGB}{0,32,96}
\definecolor{amber}{rgb}{1.0, 0.75, 0.0}
\definecolor{amaranth}{rgb}{0.9, 0.17, 0.31}

\newcount\Comments  
\newcommand{\viraj}[1]{{\ifnum\Comments=1\textcolor{green}{[viraj: #1]}\fi}}
\newcommand{\zifan}[1]{{\ifnum\Comments=1\textcolor{blue}{[zifan: #1]}\fi}}
\newcommand{\bo}[1]{{\ifnum\Comments=1\textcolor{cyan}{[bo: #1]}\fi}}
\newcommand{\commentp}[1]{{\ifnum\Comments=1\textcolor{red}{[Peter: #1]}\fi}}
\newcommand{\remove}[1]{{\ifnum\Comments=1\textcolor{red}{[#1]}\fi}}

\title{Benchmarking Massively Parallelized Multi-Task Reinforcement Learning for Robotics Tasks}

\setrunningtitle{Benchmarking Massively Parallelized Multi-Task Reinforcement Learning for Robotics Tasks}

\author{Viraj Joshi\textsuperscript{1,$\dagger$}, Zifan Xu\textsuperscript{1,$\dagger$}, Bo Liu\textsuperscript{1}, Peter Stone\textsuperscript{1,2}, Amy Zhang\textsuperscript{1}}

\emails{\{viraj\_joshi,zfxu\}@utexas.edu}

\affiliations{
$^{1}$\textbf{The University of Texas at Austin}\\
$^{2}$\textbf{Sony AI}\\
$^\dagger$ equal contribution
}

\contribution{
    This paper introduces MTBench, a unified GPU-accelerated benchmark for massively parallelized multi-task reinforcement learning (MTRL) in two robotics settings, manipulation and locomotion. 
    }
    {
    Existing robotics MTRL benchmarks, such as Meta-World \citep{yu2021metaworldbenchmarkevaluationmultitask}, have impractically long experimental runtimes, hindering the development and reproducibility of MTRL research. Other GPU-accelerated benchmarks for robotics do not support MTRL out of the box. We address both of these concerns with our end-to-end MTRL benchmark.
    }

\contribution{
    This paper conducts comprehensive experiments to evaluate all aspects of MTRL, including base RL algorithms, gradient manipulation methods, and neural network architectures.
    }
    {
    We confirm whether the reliance on off-policy methods in the MTRL literature holds in the massively parallel regime, and then evaluate a suite of MTRL schemes using on-policy methods across our evaluation settings.
    
    }

\contribution{
    This paper presents four key observations on applying existing MTRL schemes to massively parallelized training in robotics. These insights guide the selection of MTRL schemes and inform future research directions.
    }
    {
    Massively parallelized training is emerging as a popular paradigm, introducing unique challenges for existing RL methods \citep{d2022sample,li2023parallel,gallici2024simplifyingdeeptemporaldifference,sapg2024}. However, MTRL development has yet to leverage this paradigm.
    }


\summary{Multi-task Reinforcement Learning (MTRL) has emerged as a critical training paradigm for applying reinforcement learning (RL) to a set of complex real-world robotic tasks, which demands a generalizable and robust policy. At the same time, \emph{massively parallelized training} has gained popularity, not only for significantly accelerating data collection through GPU-accelerated simulation but also for enabling diverse data collection across multiple tasks by simulating heterogeneous scenes in parallel. However, existing MTRL research has largely been limited to off-policy methods like SAC in the low-parallelization regime. 
MTRL could capitalize on the higher asymptotic performance of on-policy algorithms, whose batches require data from current policy, and as a result, take advantage of massive parallelization offered by GPU-accelerated simulation. 
To bridge this gap, we introduce a massively parallelized \textbf{M}ulti-\textbf{T}ask \textbf{Bench}mark for robotics (MTBench), an open-sourced benchmark featuring a broad distribution of 50 manipulation tasks and 20 locomotion tasks, implemented using the GPU-accelerated simulator IsaacGym. MTBench also includes four base RL algorithms combined with seven state-of-the-art MTRL algorithms and architectures, providing a unified framework for evaluating their performance. Our extensive experiments highlight the superior speed of evaluating MTRL approaches using MTBench, while also uncovering unique challenges that arise from combining massive parallelism with MTRL. 
}

\begin{document}

\makeCover  
\maketitle  

\begin{abstract}
Multi-task Reinforcement Learning (MTRL) has emerged as a critical training paradigm for applying reinforcement learning (RL) to a set of complex real-world robotic tasks, which demands a generalizable and robust policy. At the same time, \emph{massively parallelized training} has gained popularity, not only for significantly accelerating data collection through GPU-accelerated simulation but also for enabling diverse data collection across multiple tasks by simulating heterogeneous scenes in parallel. However, existing MTRL research has largely been limited to off-policy methods like SAC in the low-parallelization regime. 
MTRL could capitalize on the higher asymptotic performance of on-policy algorithms, whose batches require data from the current policy, and as a result, take advantage of massive parallelization offered by GPU-accelerated simulation. 
To bridge this gap, we introduce a massively parallelized \textbf{M}ulti-\textbf{T}ask \textbf{Bench}mark for robotics (MTBench), an open-sourced benchmark featuring a broad distribution of 50 manipulation tasks and 20 locomotion tasks, implemented using the GPU-accelerated simulator IsaacGym. MTBench also includes four base RL algorithms combined with seven state-of-the-art MTRL algorithms and architectures, providing a unified framework for evaluating their performance. Our extensive experiments highlight the superior speed of evaluating MTRL approaches using MTBench, while also uncovering unique challenges that arise from combining massive parallelism with MTRL. Code is available
at \href{https://github.com/Viraj-Joshi/MTBench}{ https://github.com/Viraj-Joshi/MTBench}

\end{abstract}

\section{Introduction}
\begin{figure}[t!]
    \centering
    \includegraphics[width=\columnwidth]{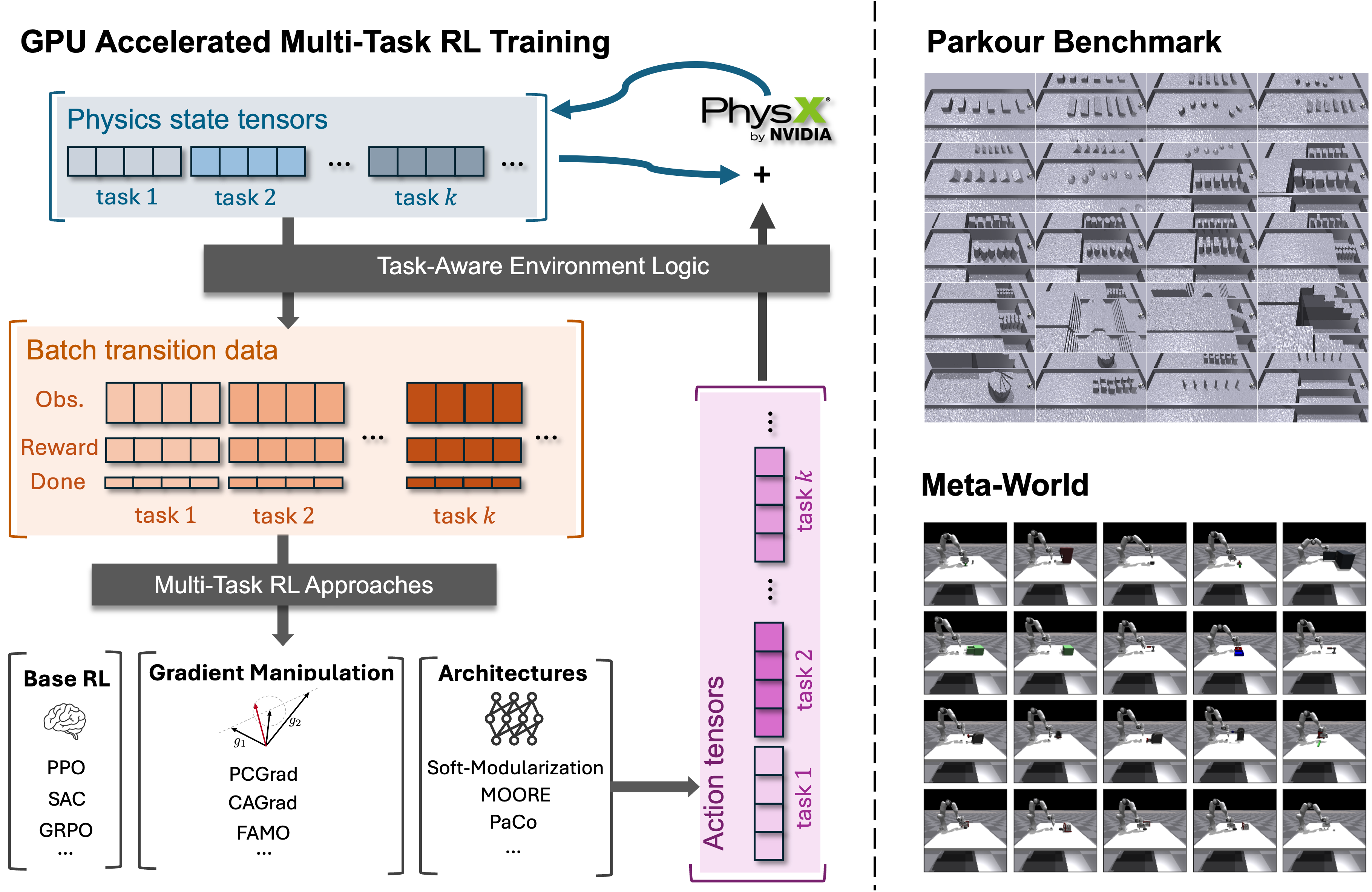}
    \caption{MTBench is a benchmark that leverages massive parallelism for MTRL in two robotics domains, Parkour and Meta-World, and provides MTRL implementations developed over the years. On the left, we see that IsaacGym's Tensor API enables us to assign blocks of environments to a desired task within the domain of interest, allowing for the setting and getting of the required information for RL training.}
    \label{fig:main}
\end{figure}
\vspace{-10pt}
Deep reinforcement learning has been successfully applied to a wide range of decision-making tasks, including Atari games \citep{mnih2013playingatarideepreinforcement}, the game of Go \citep{silver2016mastering}, and continuous control tasks \citep{hwangbo2019learning,wurman2022outracing}. While these applications have achieved remarkable task-specific performance, recent research trends have shifted towards developing general-purpose agents capable of solving multiple tasks or adapting to diverse environments \citep{cobbe2020leveraging, kirk2023survey,park2024ogbench}.
This transition is partly motivated by the demands of real-world robotics applications, where versatility and robustness are essential. For example, tabletop manipulation often requires acquiring multiple skills to accomplish complex tasks \citep{pinto2016learningpushgraspingusing, yu2021metaworldbenchmarkevaluationmultitask} and legged locomotion demands adaptability to traverse challenging terrains \citep{lee2020learning, liang2024eurekaverseenvironmentcurriculumgeneration}.

To facilitate the learning of a general-purpose robotic agent, massively parallelized training (>>1000 simulations) has gained popularity with the advancement of GPU-accelerated simulators \citep{liang2018gpu,freeman2021brax,makoviychuk2021isaacgymhighperformance,mittal2023orbit, taomaniskill3, mujoco_playground_2025}. These simulators have significantly mitigated hardware and runtime constraints for learning \emph{single tasks}, reducing experiment durations from days to minutes \citep{liang2018gpu,rudin2022learning}. However, in the multi-task setting, no out-of-the-box solution exists to allocate a fixed number of environments per task on a single GPU, allowing for simultaneous diverse data collection and end-to-end MTRL training. 
Additionally, massively parallelized online batched RL introduces new, non-trivial algorithmic challenges. For example, on-policy methods like PPO reach a saturation point beyond which additional parallelization no longer improves performance \citep{sapg2024}. Meanwhile, off-policy methods such as SAC and Q-Learning become unstable, losing their sample efficiency compared to on-policy methods as interaction with parallel environments unbalances the replay ratio \citep{d2022sample,li2023parallel,gallici2024simplifyingdeeptemporaldifference}.

On the other hand, learning general-purpose robotic agents has also motivated multi-task RL (MTRL), which aims to learn a single policy that maximizes average performance across multiple tasks. By leveraging task similarities \citep{pinto2016learningpushgraspingusing}, MTRL often enhances sample efficiency, requiring fewer transitions to match the performance of single-task counterparts. Prior research has primarily focused on addressing optimization challenges introduced by multiple learning signals, either from a gradient-based perspective \citep{yu2020gradientsurgerymultitasklearning, liu2024conflictaversegradientdescentmultitask, liu2023famofastadaptivemultitask} or through neural architecture design \citep{yang2020multitaskreinforcementlearningsoft, Sodhani2021MultiTaskRL, Sun2022PaCoPM, hendawy2024multitaskreinforcementlearningmixture}. However, these MTRL approaches have been limited to using off-policy methods in low-parallelization settings using libraries like Ray \citep{liang2018rllib}. With massive parallelization applied to MTRL, we no longer need to deal with how to distribute experience collection and learning, instead utilizing on-policy algorithms, whose batches require data from current experience and as a result, take advantage of the parallelization offered by GPU-based simulators.



To support large-scale MTRL experiments and advance the development of general-purpose robotic agents, we introduce a massively parallelized \textbf{M}ulti-\textbf{T}ask \textbf{Bench}mark for robotics (MTBench). This open-source benchmark includes a diverse set of 50 manipulation tasks and 20 locomotion tasks (right side of Figure \ref{fig:main}), implemented using the GPU-accelerated simulator IsaacGym. Each task allows for procedurally generating infinitely many variations by modifying factors such as initial states and terrain configurations. Additionally, MTBench integrates four base RL algorithms with seven state-of-the-art MTRL algorithms and architectures, providing a unified framework to evaluate their performance.



Based on our experiments, we highlight the following major observations:
\vspace{-10pt}
\paragraph{(O1) On-Policy > Off-Policy:} Choosing between on-policy RL methods or off-policy methods affects performance more than the MTRL scheme applied in massively parallel training. Off-policy RL's asymptotic performance struggles to match on-policy RL in this regime.
\vspace{-10pt}
\paragraph{(O2) Prioritize Wall-Clock Time over Sample Efficiency:} In the massively parallel regime, wall-clock efficiency is more critical than sample efficiency, as experience collection scales easily with more GPUs. 
\vspace{-10pt}
\paragraph{(O3) Value Learning is the Key Bottleneck in MTRL:} Multi-task RL struggles primarily with value estimation rather than policy learning, as gradient conflicts mostly impact the critic function. 
\vspace{-10pt}
\paragraph{(O4) Curriculum Learning is Crucial for Sparse-Reward Tasks:} MTRL alone does not help exploration in sparse-reward tasks; curriculum learning is essential for overcoming early stagnation. 

\section{Background}
\subsection{GPU Accelerated Simulation}
Traditionally, simulators used for online RL rely on the coordination between CPU and GPU where the CPU handles physics simulation and observation/reward calculations while the GPU handles neural network training and inference, leading to frequent slow memory transfers between the two many times during the RL training process.  Now, GPU-accelerated simulators provide access to the results of physics simulation on the GPU, and as a result, we have all relevant data - observations, actions, and rewards - remaining on the GPU throughout the learning process. This development allows for massive parallelization and as a result, dramatically reduces MTRL training time from days or weeks on thousands of CPU cores to just hours on a single GPU. 


Specifically, NVIDIA IsaacGym offers a Tensor API that directly exposes the physics state of the world in Python, so we can directly populate and manage massively parallelized heterogeneous scenes for all tasks (Figure \ref{fig:main}), avoiding the communication overhead of synchronizing experience collection and neural network training across distributed systems \citep{nair2015massively,espeholt2018impala}.

\subsection{Multi-Task Reinforcement Learning}

RL is formalized as a finite horizon, discrete-time MDP, which is represented by a tuple $\mathcal M = (\mathcal S, \mathcal A, \mathcal P, r, \mu, \gamma)$, where $\mathcal S \in \mathbb R^n$ denotes the continuous state space, $\mathcal A \in \mathbb R^m$ denotes the continuous action space, $\mathcal P: \mathcal S \times \mathcal A \rightarrow \Delta(\mathcal S)$ denotes the stochastic transition dynamics, $r: \mathcal S \times \mathcal A \rightarrow \mathbb R$ denotes the reward function, $\mu: \mathcal S \rightarrow \Delta(\mathcal S)$ denotes the initial state distribution, and $\gamma \in [0,1)$ is the discount factor. A policy parameterized by $\theta$, $\pi_\theta(a_t|s_t): \mathcal{S} \rightarrow \Delta(\mathcal A)$, is a probability distribution over actions conditioned on the current state. RL learns a policy $\pi_\theta$ such that it maximizes the expected cumulative discounted return $J(\theta) = \mathbb E_{s_0 \sim \mu,\pi_\theta} [\sum_{t=0}^T \gamma^t r(s_t,a_t)]$ where $a_t \sim \pi_\theta$. 

\paragraph{Problem statement}
Each task $\tau$ is sampled the task distribution $p(\mathcal T)$ is a different MDP $\mathcal M^\tau = (\mathcal S^\tau, \mathcal A^\tau, \mathcal P^\tau, r^\tau, \mu^\tau, \gamma^\tau)$. MTRL learns a single policy $\pi_\theta$ that maximizes the expected cumulative discounted return averaged across all tasks $J(\theta) = \sum_{\tau \in \mathcal T} J_\tau (\theta)$. The only restriction we place upon $\mathcal M^\tau$ is that their union shares a universal state space $\mathcal S$ and by appending a task embedding $z$ to the state, we give the policy the ability to distinguish what task each observation belongs to.


A change in any part of a $\mathcal M^\tau$ constitutes what it means to define a new task. In locomotion, each task from $p(\mathcal T)$ would be associated with a different goal to reach \textit{in the same control setting}, so only $r^\tau$ would differ across tasks. In tabletop manipulation like Meta-World, the tasks range from basic skills like pushing and grasping to more advanced skills combining these basic skills, so the goals ($r^\tau$) and state spaces ($\mathcal S^\tau$) vary across tasks but the action spaces $A^\tau$ are identical.

\section{Benchmark}
MTBench provides a unified framework for simulating two key robotics task categories: manipulation and locomotion, within the IsaacGym simulator. For manipulation, we incorporate 50 tasks from Meta-World \citep{yu2021metaworldbenchmarkevaluationmultitask}, chosen for their simplicity, task diversity, and well-designed, shaped rewards. The locomotion domain includes 20 diverse quadrupedal Parkour tasks from Eurekaverse \citep{liang2024eurekaverseenvironmentcurriculumgeneration}, the most comprehensive Parkour benchmark, encompassing a wide range of established locomotion challenges. As Figure \ref{fig:main} demonstrates, MTBench supports defining any custom subset of tasks and their associated number of environments, enabling researchers to craft different task sets of varying difficulty. This section provides a detailed overview of these task domains and the evaluation protocols.

\subsection{Meta-World}
\begin{figure}[htb!]
    \centering
    \includegraphics[width=0.95\linewidth]{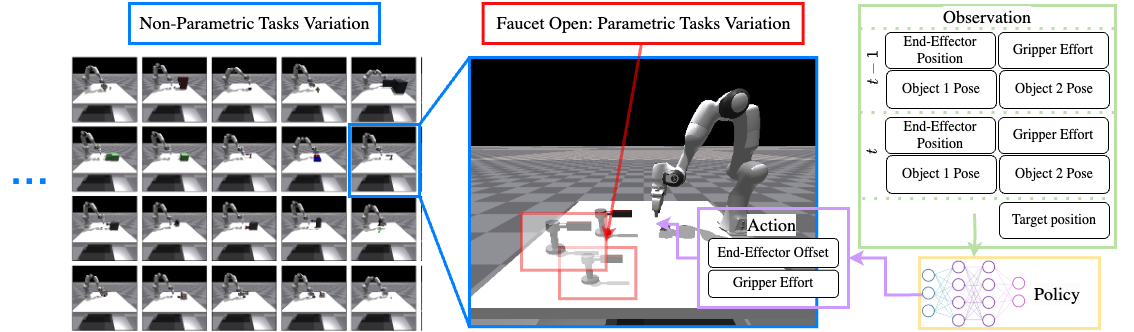}
    \caption{Illustrations of non-parametric tasks variation, parametric tasks variation of Faucet Open, and the observation and action space of the RL agents in the Meta-World benchmark.}
    \label{fig:metaworld_viz}
\end{figure}
\paragraph{Task Descriptions: } Meta-World consists of 50 tabletop manipulation tasks that require a simulated one-armed robot \citep{FrankaPanda} to interact with one or two objects in various ways, such as pushing, picking, and placing. Within each task, Meta-World provides parametric variation over the initial object position and target position. Each task has a pre-defined success criterion. Our re-implementation of Meta-World makes necessary changes by updating Sawyer to Franka Emika Panda and tuning the reward function of each task to ensure that the tasks are individually solvable.

\paragraph{Observation and Action Spaces:}  Despite sharing a common state space dimensionality, the semantic meaning of certain dimensions varies across tasks. The state representation comprises the end-effector's 3D position in $\mathbb{R}^3$, the normalized gripper effort in $\mathbb{R}^1$, the object 3D positions from two objects in $\mathbb{R}^6$, and the quaternion representation of the two objects' orientation in $\mathbb{R}^8$. For tasks involving a single object, the state dimensions corresponding to a second object are set to zero. To account for temporal dependencies, the observation space concatenates the state representations from two consecutive time steps and appends the 3D position of the target goal. This results in a final observation vector of 39 dimensions. The action space is also consistent across the tasks, comprising of the displacement of the end-effector in $\mathbb{R}^3$ and the normalized gripper effort in $\mathbb{R}^1$. An overview of the observation and action can be seen in Figure \ref{fig:metaworld_viz}.

\paragraph{Evaluation Settings:} Following \citet{yu2021metaworldbenchmarkevaluationmultitask}, we explictly provide two evaluation settings: multi-task 10 (MT10) and multi-task 50 (MT50), where MT10 consists of 10 selected tasks and MT50 consists of all 50 tasks. During evaluation, we measure the \emph{success rates} (SR) (Appendix \ref{appendix:meta-world-success}) and the \emph{cumulative reward} (R). When each \textit{environment} has its parametric parameters randomly varied every reset, the evaluation is referred to as MT10-rand and MT50-rand.

\subsection{Parkour Benchmark}
\begin{figure}[htb!]
    \centering
    \includegraphics[width=0.9\linewidth]{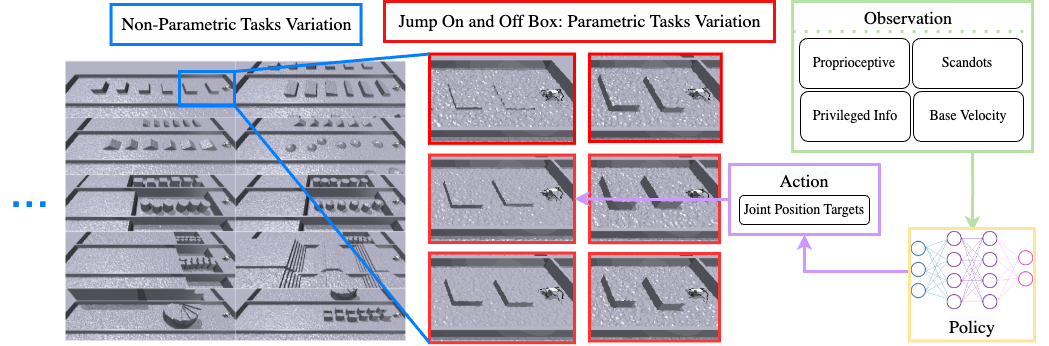}
    \caption{Illustrations of non-parametric tasks variation, parametric tasks variation of Jump On and Off Box, and the observation and action space of the RL agents in the Parkour benchmark.}
    \label{fig:parkour_viz}
\end{figure}
\paragraph{Task Descriptions:} In the Parkour tasks from Eurekaverse, the agent controls a quadrupedal Unitree Go1 robot \citep{unitree_go1} to track predefined waypoints while traversing one of 20 different terrain categories \citep{liang2024eurekaverseenvironmentcurriculumgeneration}. These tasks challenge various motor skills, including climbing boxes, walking on slopes, jumping, navigating stepping stones, ascending stairs, maneuvering through narrow hallways, weaving through agility poles in a zig-zag pattern, and maintaining balance. The left side of Figure \ref{fig:parkour_viz} shows bird-eye views of these terrain categories.

Each task also provides parametric terrain variations defined by a set of terrain parameters, whose definitions and valid ranges are detailed in the supplementary materials. Additionally, each task introduces a one-dimensional continuous variable, termed \emph{difficulty}, and a predefined mapping from the \emph{difficulty} to a set of terrain parameters. This \emph{difficulty} measure aligns with human intuition; for instance, high boxes present a greater challenge than lower boxes for a quadrupedal robot to jump on and off.

\paragraph{Observation and Action Spaces:} The observation of the agent is slightly simplified for more efficient benchmarking compared to Eurekaverse. The observation is compromised by proprioceptive observation in $\mathbb{R}^{48}$, scandots of the terrain environments in $\mathbb{R}^{132}$, base linear velocity in $\mathbb{R}^{3}$, and privileged information in $\mathbb{R}^{29}$. The action assigns joint position targets at a frequency of 50 Hz for a Proportional-Derivative (PD) controller. An overview of the observation and actions is shown in Figure \ref{fig:parkour_viz}. The reward function resembles \citet{fu2023deep}, which encourages positive linear and angular velocities that point to the next waypoint, while minimizing energy consumption. 

\paragraph{Evaluation Settings:} We define two evaluation settings: Parkour-easy and Parkour-hard. Parkour-easy consists of 200 terrains, with each of the 20 tasks assigned 10 terrains generated at the lowest difficulty level. In contrast, Parkour-hard also includes 200 terrains but distributes difficulty levels uniformly across the 10 terrains per task, providing a more diverse and challenging evaluation setting. Before the training, all the evaluated methods are pre-trained on flat ground to acquire the basic walking gait. Such a pre-training phase is typical in the literature \citep{zhuang2023robot,cheng2024extreme}.

During evaluation, we measure \emph{progress} (P) as the ratio of the current waypoint index to the total number of waypoints at the time of episode termination. An agent that successfully traverses the entire terrain achieves a \emph{progress} score of 100\%. The overall \emph{progress} is computed as the average over 200 terrains, with each terrain evaluated across 10 independent runs.
\begin{figure}[t!]          
    \includegraphics[width=.95\linewidth]{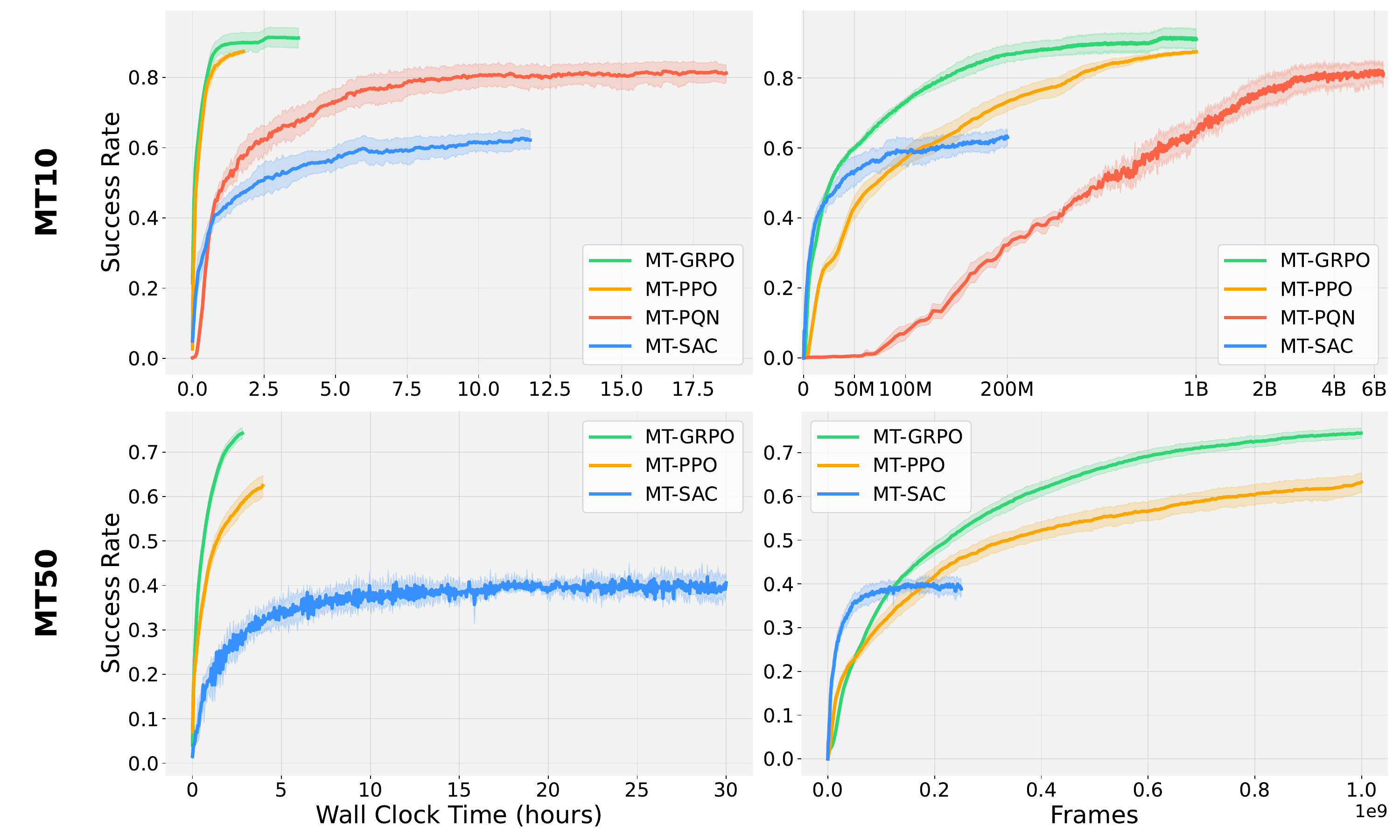}
    \caption{\textbf{Vanilla MTRL performance in Meta-World.} We report the pointwise 95\% percentile bootstrap CIs of the average success rates using 10 seeds for each RL algorithm in the MT10-rand and MT50-rand evaluation settings. On-policy methods (MT-PPO, MT-GRPO) continue to improve with more experience, achieving a substantially higher success rate than the traditional off-policy method, MT-SAC, in substantially less time.}
    \label{fig:efficiency}
\end{figure}
\subsection{Algorithms}
We re-implement a suite of algorithms and MTRL approaches using a popular learning library RLGames \citep{rl-games2021}, providing a unified benchmark for end-to-end vectorized MTRL training across many seeds and hyperparameters on a single GPU. Our benchmark is highly extensible towards new RL algorithms as well as approaches within the two axes of MTRL research, gradient manipulation, and neural architectures. There is a brief overview in Appendix \ref{appendix:mtrl_overview}. 

\paragraph{Base MTRL Algorithms}
We implement four RL algorithms: MT-PPO, a multi-task version of Proximal Policy Optimization \citep{schulman2017proximalpolicyoptimizationalgorithms}; MT-GRPO \citep{shao2024deepseekmathpushinglimitsmathematical}, a variant of PPO introduced for language modeling but adapted here for control; MT-SAC, a multi-task version of Soft Actor-Critic \citep{haarnoja2018softactorcriticoffpolicymaximum}; and MT-PQN, a novel multi-task extension to Parallel Q-learning \citep{gallici2024simplifyingdeeptemporaldifference} to handle continuous control problems. All algorithms are multi-task versions of their single-task counterparts, simply by augmenting the observation space with one-hot task embeddings.

\paragraph{MTRL Schemes}
We implement two categories of MTRL schemes that can be easily combined with any of our base algorithms. The first category consists of gradient manipulation methods: PCGrad \citep{yu2020gradientsurgerymultitasklearning}, CAGrad \citep{liu2024conflictaversegradientdescentmultitask}, and FAMO \citep{liu2023famofastadaptivemultitask}. The second category consists of multi-task architectures: Soft-Modularization \citep{yang2020multitaskreinforcementlearningsoft}, CARE \citep{Sodhani2021MultiTaskRL}, PaCo \citep{Sun2022PaCoPM}, and MOORE \citep{hendawy2024multitaskreinforcementlearningmixture}. The prefix "MH" (multi-head) is prepended to name of the MTRL approach to denote one output head per task, and otherwise "SH" (single-head) to denote tasks sharing one head.

\paragraph{Curriculum Learning}
Unlike Meta-World, where reward functions are carefully designed with dense rewards, locomotion tasks often rely on sparse reward signals (e.g., moving forward to the next waypoints). As a result, strategies like curriculum learning have been widely adopted to facilitate learning in challenging tasks, such as running \citep{margolis2024rapid} and jumping onto high platforms \citep{liang2024eurekaverseenvironmentcurriculumgeneration}. Inspired by this, we incorporate a simple curriculum strategy to train Parkour-hard tasks. In Parkour-hard, each task consists of ten terrains with varying levels of \emph{difficulty}. Agents always begin on the easiest terrain and progress to more challenging ones if they achieve a \emph{progress} of at least 80\%  in their current terrain. 
We refer to the Parkour-hard training with curriculum learning by Parkour-hard-cl.
\section{Results}
\begin{figure}[t!]          
    \includegraphics[width=.95\linewidth]{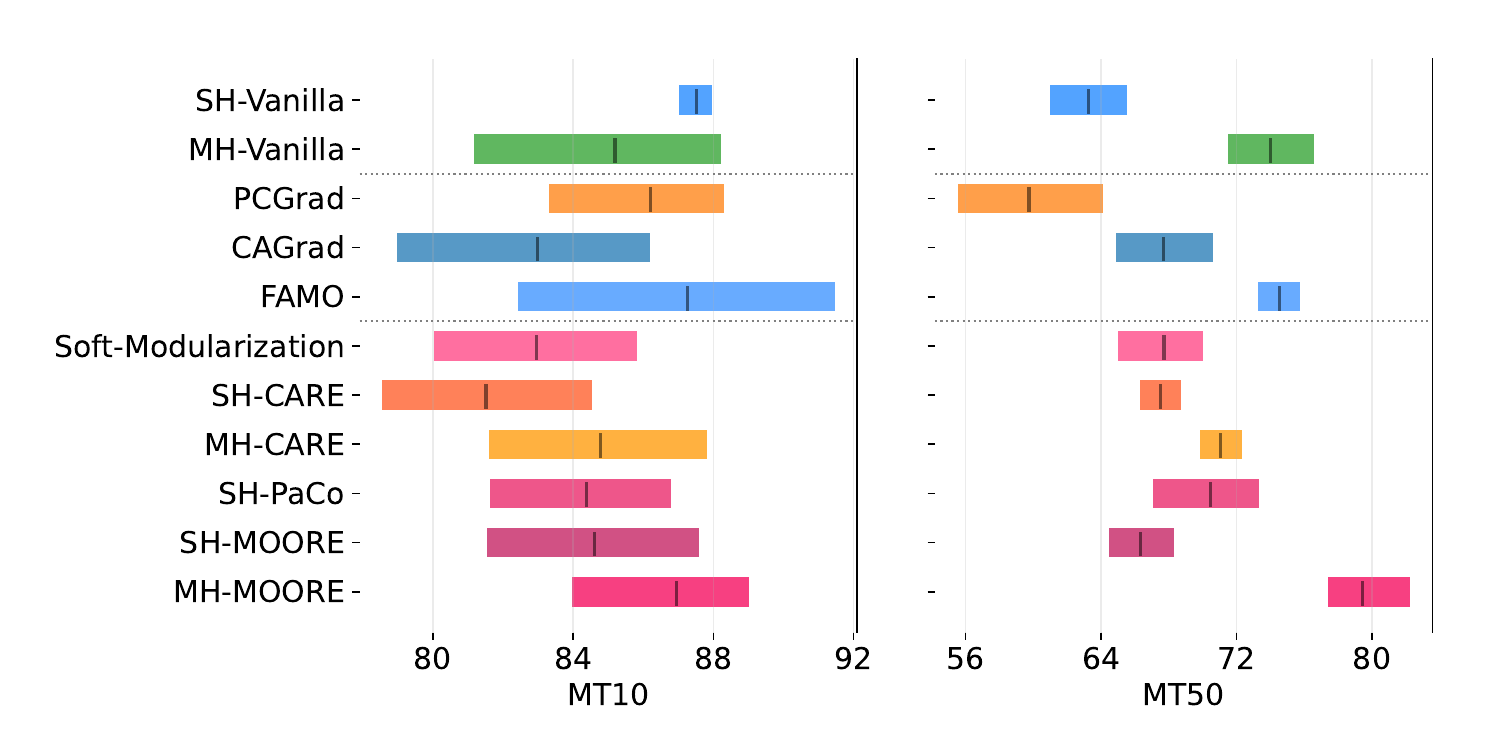}
    \caption{We compare the 95\% bootstrapped confidence intervals of the average success rate of all MTRL approaches using MT-PPO for the MT10-rand and MT50-rand evaluation settings of Meta-World. Each approach uses 1B frames per run over 10 seeds. Exact numbers are in Table \ref{table:main_table_results_metaworld}.}
    \label{fig: main_result_metaworld}
\end{figure}
In this section, we present the results of our benchmark across our evaluation settings and empirically justify the aforementioned four major observations. 

\subsection{Choosing the MTRL Algorithm (\textbf{O1, O2})}
To illustrate how on-policy MTRL methods leverage massive parallelism, we first evaluate two on-policy methods, MT-PPO and MT-GRPO, alongside two off-policy methods, MT-SAC and MT-PQN, in Meta-World. Figure \ref{fig:efficiency} presents the learning curves with respect to both wall-clock time and the number of environment interactions. Since this observation is concerned with answering what the best base MTRL algorithm is, we tune all aspects of each method to achieve its highest success rate, including using different network architectures. The full hyperparameter and model details are in the supplementary materials.

\paragraph{On-policy methods outperform traditional off-policy methods.}

Using MT-SAC as a representative of traditional off-policy algorithms used for MTRL, Figure \ref{fig:efficiency} shows there is a substantial performance gap in success rates between MT-PPO and MT-SAC in both evaluation settings and, more importantly, a substantial wall-clock time difference as well (roughly 22 minutes and 12 hours after 200M frames of collected experience in MT10-rand). While MT-SAC can match MT-PPO's runtime by simply matching the gradient steps per epoch that MT-PPO takes, this results in a near-zero success rate. Furthermore, as the number of tasks increases to the MT50-rand setting, these gaps increase. 

Traditional off-policy methods in the massively parallelized regime cannot effectively leverage increased environment interaction, as their stability, performance, and runtime greatly rely on the ratio of gradient updates to environment steps, i.e, update-to-data (UTD) ratio \citep{d2022sample} being greater than or equal to 1.  Research into leveraging massive parallelism in off-policy methods is gaining popularity but is either not yet adapted for the continuous control setting \citep{gallici2024simplifyingdeeptemporaldifference} or requires distributed asynchronous processes spread across GPUs \citep{li2023parallel}.



\paragraph{Off-Policy methods can be designed for the massively parallelized regime.} In Figure \ref{fig:efficiency}, we also included our adaptation of PQN \citep{gallici2024simplifyingdeeptemporaldifference} to the multi-task continuous control setting. The details of our implementation are in  Appendix \ref{appendix:PQN}. Surprisingly, applying these simple changes to an originally discrete control algorithm and left to run long enough, MT-PQN can roughly match the performance of MT-PPO in MT10-rand. Considering PQN's performance and stability, similar simulation throughput to PPO, and lack of a replay buffer, suggests that smartly adapting PQN to multi-task continuous control tasks could be a promising research direction compared to actor-critic algorithms.

\begin{figure}[t!]
\centering
\begin{minipage}{.5\textwidth}
  \centering
  \includegraphics[width=.95\linewidth]{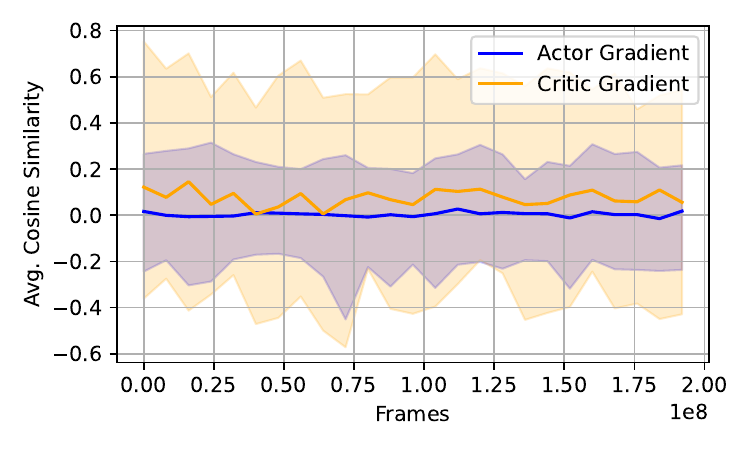}
  \captionof{figure}{The average gradient cosine similarity across all task pairs in MT10-rand for both actor and critic networks. The shadow areas represent the ranges between minimum and maximum cosine similarities.}
  \label{fig:cosine_sim}
\end{minipage}%
\quad
\begin{minipage}{.45\textwidth}
  \centering
  \includegraphics[width=0.85\linewidth]{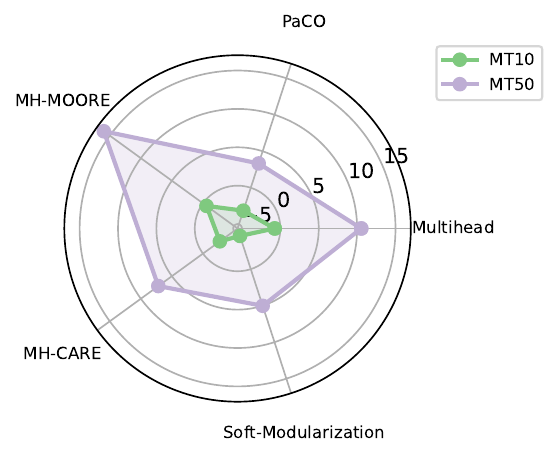}
  \captionof{figure}{Success rate (SR) differences of five neural network architectures relative to the Vanilla baseline in \textcolor[HTML]{80c97f}{MT10-rand} and \textcolor[HTML]{beaed4}{MT50-rand}.}
  \label{fig:lidar_nn}
\end{minipage}
\end{figure}

\subsection{MTRL Approaches (\textbf{O2, O3})} Figure \ref{fig: main_result_metaworld} reports the 95\% bootstrap confidence intervals of the mean success rate following \citet{agarwal2021deep} of all MTRL approaches using MT-PPO. All of the gradient manipulation methods use the same three-layer MLP neural networks. 


\paragraph{Multi-task architectures show greater performance gains with larger task sets.} As shown in Figure \ref{fig:lidar_nn}, the benefits of multi-task architectures become more pronounced as the number of tasks increases. In MT10-rand, vanilla PPO asymptotically outperforms advanced multi-task architectures. However, in MT50-rand, the best-performing multi-task architecture, MH-MOORE, surpasses the vanilla approach by roughly 16\% in success rate. This improvement is likely due to enhanced knowledge sharing that only manifests in training diverse enough tasks, such as MT50. However, similar performance gains are not observed in the Parkour benchmark, likely due to the insufficient task diversity in Parkour tasks.

\begin{figure}[t!] 
    \begin{minipage}[c]{0.49\columnwidth} 
        \centering
        \footnotesize 
        \resizebox{\linewidth}{!}{
        \begin{tabular}{lccc}
        \toprule
        \multirow{2}{*}{\diagbox{Methods}{Tasks}} & Parkour-easy & Parkour-hard & Parkour-hard-CL \\
        \cmidrule(lr){2-2} \cmidrule(lr){3-3} \cmidrule(lr){4-4}
        & P(\%) $\uparrow$ & P(\%) $\uparrow$ & P(\%) $\uparrow$ \\
        \midrule
        Vanilla & $80.39 \pm 0.43$ & $54.51 \pm 0.82$ & $68.12 \pm 0.43$\\
        Multihead & $73.17 \pm 2.53$ & $49.65 \pm 1.27$ & $62.17 \pm 1.05$ \\
        \midrule
        PCGrad & \textbf{80.61 $\pm$ 0.78} & \textbf{55.98 $\pm$ 0.24} & $68.37 \pm 0.65$ \\
        CAGrad & $80.25 \pm 0.32$ & $55.88 \pm 0.91$ & $67.75 \pm 0.43$\\
        FAMO & $79.79 \pm 0.29$ & $55.56 \pm 0.97$ & \textbf{68.57 $\pm$ 0.76}\\
        \midrule
        PaCo & $78.63 \pm 0.70$ & \textbf{58.65 $\pm$ 0.77} & $64.15 \pm 0.98$\\
        SH-MOORE & $64.61 \pm 1.69$ & $46.78 \pm 0.65$ & $49.53 \pm 0.52$\\
        Soft-Modularization & $69.29 \pm 3.81$ & $47.28 \pm 0.23$ & $51.43 \pm 0.35$ \\
        \bottomrule
        \end{tabular}
        }
        \captionof{table}{The average success rate and standard deviation of MTRL approaches using MT-PPO in all Parkour evaluation settings. Each approach uses 250M frames per run over 10 seeds.}
        \label{table:main_result_parkour}
    \end{minipage}
    \hfill 
    \begin{minipage}[c]{0.49\columnwidth} 
        \centering
        \footnotesize 
        \includegraphics[width=0.7\linewidth]{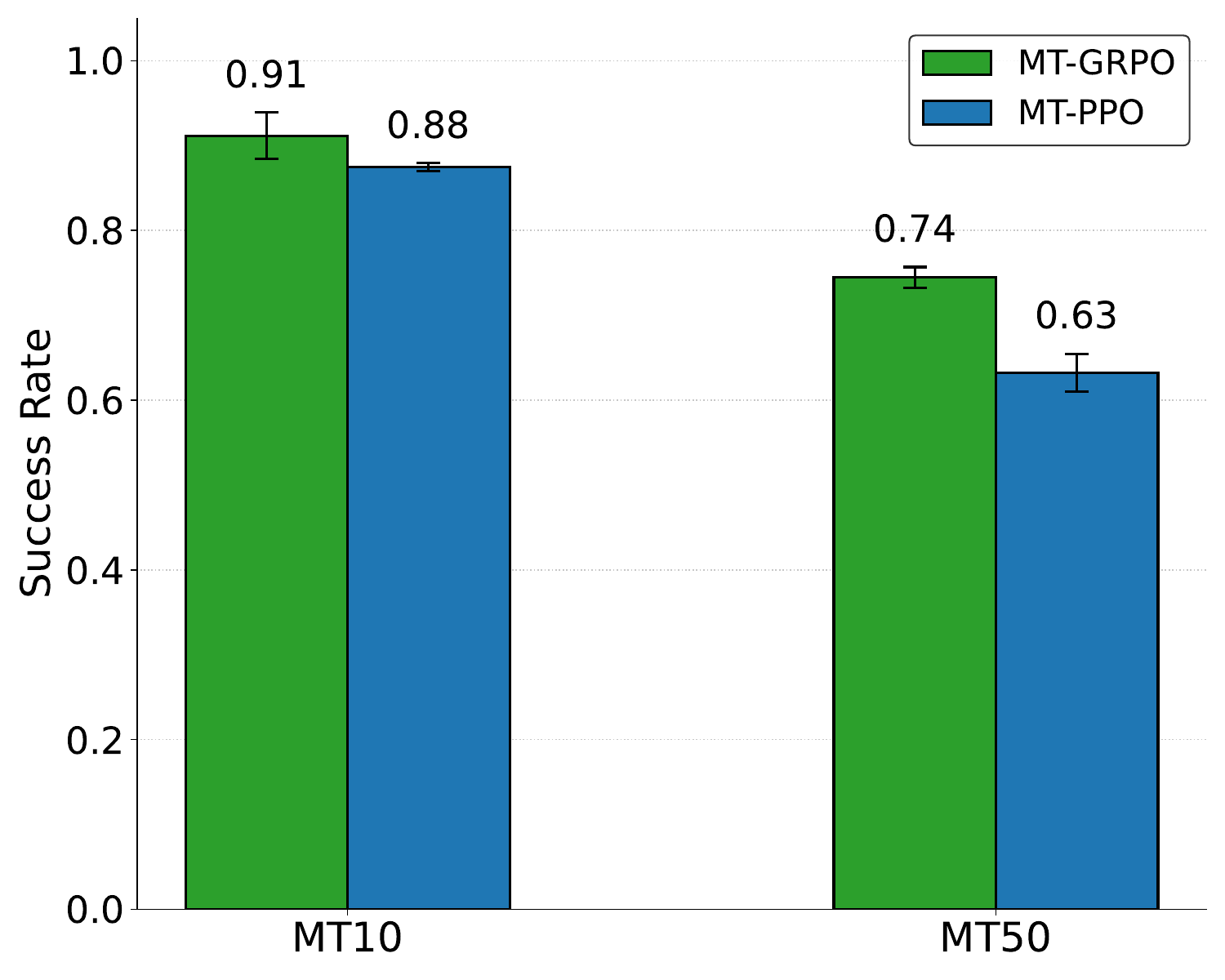}
        \captionof{figure}{\textbf{Eliminating the difficulty of critic estimation consistently improves performance over most MTRL approaches using MT-PPO} when comparing the 95\% bootstrapped CI of average success rates in Meta-World. Each approach uses 1B frames per run over 10 seeds. }
        \label{fig:grpo_vs_vanilla}
    \end{minipage}
\end{figure}

\paragraph{Resolving gradient conflict consistently improves the performance.} Gradient manipulation can outperform or match vanilla MT-PPO across all evaluation settings (middle section of Figure \ref{fig: main_result_metaworld} and Table \ref{table:main_result_parkour}). This suggests that gradient conflicts are still a common optimization challenge in multi-task RL problems. Among these methods, FAMO shows superior scalability with respect to an increasing number of tasks in its success rate as well as wall-clock training time, likely due to its simple strategy of adaptive task weighting, which eliminates the need for backpropagating through each task's loss.

\paragraph{Value learning is the key bottleneck in MTRL. } Prior research in MTRL has shown that addressing gradient conflicts improves performance in off-policy actor-critic RL algorithms like MT-SAC. Our benchmarking results extend this observation to on-policy actor-critic algorithms, demonstrating that gradient conflicts also arise when learning the critic network in MT-PPO. However, we do not observe similar conflicts in policy optimization. This observation aligns with prior work using an actor-critic algorithm for large-scale multi-task learning \citep{hessel2019multi}. Figure \ref{fig:cosine_sim} shows the average cosine similarity across all task gradient pairs for both actor and critic networks, where critic gradients manifest lower minimum similarities.

\subsection{Reward Sparsity (O4)}
Although tasks in Meta-World and the Parkour Benchmark are defined independently of their reward functions, training performance is significantly influenced by reward design. We adopt commonly used reward formulations in both domains. In Meta-World, tasks utilize dense rewards, which provide continuous feedback to guide specific interactions between the robotic arm and objects. In contrast, the Parkour Benchmark employs a sparse reward scheme, where the agent is rewarded solely for maintaining forward velocity toward waypoints, without receiving additional signals for intermediate behaviors. 

\paragraph{Dense rewards increase the complexity of multi-task critic learning.} In multi-task RL, dense reward functions introduce challenges for critic learning, as different tasks exhibit varying reward distributions and gradient magnitudes. We can see in Figure \ref{fig: main_result_metaworld} that addressing these conflicts in dense-reward multi-task settings such as MT10-rand and MT50-rand can improve performance. However, the performance gains are relatively marginal in a sparse-reward multi-task setting like the Parkour benchmark. 

\paragraph{Curriculum learning is crucial for sparse-reward tasks.} In environments with sparse rewards, standard MTRL methods do not inherently enhance exploration, as agents receive limited feedback in each task. This challenge is particularly evident in the Parkour Benchmark, where agents tend to adopt overly conservative behaviors in more difficult tasks. Curriculum learning addresses this issue by structuring task progression, enabling agents to first master simpler behaviors before tackling more complex ones. By gradually increasing task difficulty, curriculum learning improves exploration efficiency and yields a 10\% performance gain in \emph{progress}, as observed when comparing Parkour-hard and Parkour-hard-cl (columns 2 and 3 in Table \ref{table:main_result_parkour}).

\subsection{Learning without a Critic (O3)}
To further investigate the impact of gradient conflict in the critic on MTRL, we can eliminate the critic by increasing the horizon length in MT-PPO to be equal to the length of the episode. 
\paragraph{MTRL can benefit from eliminating gradient conflict in the critic.} In fact, eliminating the critic from MT-PPO is equivalent to implementing MT-GRPO \citep{shao2024deepseekmathpushinglimitsmathematical} without the KL term. We use the Monte Carlo estimate of the return as the reward in the advantage calculation. In the dense reward setting, Figure \ref{fig:grpo_vs_vanilla} indicates that MT-GRPO is a simple baseline that nearly outperforms every MTRL approach (except MH-MOORE and FAMO in MT-50) using the same hyperparameters as MT-PPO and no baked-in MTRL design. 



\paragraph{Massive Parallelism is well suited for reducing bias from an imperfect critic.} By directly using Monte Carlo returns instead of bootstrapping, we effectively eliminate the bias introduced by imperfect critic estimation. This approach represents a clear bias-variance tradeoff: while removing the critic increases the variance of our gradient estimates, this increased variance can be effectively mitigated through large batches (of size episode length times the number of parallel environments) made possible by massive parallelization \citep{sutton1999policy}.
\section{Related}

\subsection{Parallelizing RL}

As deep online RL relies on training neural networks (learners) and collecting experience (actors), many methods have explored how to parallelize both aspects to speed up training over the years. Early works leveraged low levels of parallelization without hardware accelerators mainly for Atari either in a distributed compute cluster of hundreds (in some cases thousands) of CPU cores \citep{nair2015massively} or a single machine using a multi-threaded approach \citep{mnih2016asynchronous}. Hybrid CPU-GPU distributed frameworks introduce accelerating learners with GPUs \citep{babaeizadeh2016reinforcement,espeholt2018impala,horgan2018distributedprioritizedexperiencereplay,petrenko2020sample} along with actors collecting experience across CPUs.


Unlike Atari, robotic control tasks rely on physics simulators \citep{Mujoco}, where distributed RL methods using CPU-based simulators would demand even more intense hardware requirements \citep{liang2018rllib,openai2019learningdexterousinhandmanipulation} due to running multiple simulator instances in parallel. A wave of recent GPU-accelerated simulators \citep{liang2018gpu,freeman2021brax, makoviychuk2021isaacgymhighperformance, mittal2023orbit, taomaniskill3, mujoco_playground_2025} has essentially alleviated the experience collection constraint and shown success in rapidly learning single-task robotic control tasks \citep{allshire2021transferring, rudin2022learning} with the modest hardware requirement of 1 GPU. 

\subsection{GPU-Accelerated Benchmarks}
Several RL benchmarks have arisen as a result of GPU-accelerated simulation, mainly in JAX-based game environments \citep{cobbe2020leveraging,gymnax2022github,morad2023popgym,bonnet2023jumanji,koyamada2023pgx,rutherford2024jaxmarlmultiagentrlenvironments,matthews2024craftaxlightningfastbenchmarkopenended}. In contrast, a relatively small number of rigid-body robotic tasks are bundled with GPU-accelerated simulators or soft-body robotic tasks with other simulation platforms \citep{chen2022daxbench,xing2024stabilizing}. To truly represent the multi-task challenge, MTBench precludes adapting popular, small task sets, e.g, robot tasks from DMControl \citep{tassa2018deepmindcontrolsuite} or robosuite \citep{zhu2020robosuite}, or combining them since their tasks significantly overlap, resulting in low diversity.
For manipulation, Meta-World resolves both of these concerns and maintains continuity of MTRL research over other large task set alternatives like RLBench \citep{james2019rlbench} or LIBERO \citep{liu2023liberobenchmarkingknowledgetransfer}. 

In the domain of locomotion, massively parallelized training has become the standard approach due to its simplicity and increased robustness \citep{hwangbo2019learning,lee2020learning}. Nevertheless, parkour-style locomotion—which requires qualitatively different motor skills across terrains—remains a challenging setting for multi-task learning, which recent work addresses by learning specialized policies for individual motor skills and subsequently distilling them into a unified policy \citep{zhuang2023robot}. Emerging locomotion benchmarks such as HumanoidBench~\citep{sferrazza2024humanoidbench} and the Parkour Benchmark~\citep{liang2024eurekaverseenvironmentcurriculumgeneration} feature a broad variety of Parkour tasks, but have not yet been adopted for evaluating multi-task RL methods. Another important class of locomotion tasks involves humanoid motion imitation, such as those in PHC~\citep{luo2023perpetual} and LocoMuJoCo~\citep{alhafez2023b}, which exhibit inherently diverse task distributions due to the complex and high-dimensional nature of human motion.


\section{Conclusions}

We present MTBench, a highly extensible MTRL benchmark that includes a GPU-accelerated implementation of Meta-World and Parkour tasks, extensive gradient manipulation and neural architecture baselines, and an initial study on the current state as well as future directions of MTRL in the massively parallel regime. However, MTBench is limited to state-based MTRL to retain high simulation throughput, which we hope to resolve with pixel-based MTRL using NVIDIA IsaacLab in a future release of MTBench.

Future work can use MTBench beyond online MTRL methods. One can explore offline RL, imitation learning, or distillation methods by writing additional code to rapidly collect transitions from expert single-task agents. Another application of our benchmark could be as part of the ‘finetune’ step in the ‘pretrain, then finetune’ paradigm where one pre-trains on a diverse set of tasks using offline data and rapidly finetunes an agent online using our environments.

\appendix

\section{PQN}
\label{appendix:PQN}
Parallel Q-learning \citep{gallici2024simplifyingdeeptemporaldifference} is a recent off-policy TD method designed for discrete action spaces and massively parallelized GPU-based simulators that casts aside the tricks introduced over the years to stabilize deep Q learning such as replay buffers \citep{mnih2013playingatarideepreinforcement}, target networks \citep{Mnih2015HumanlevelCT} and double Q-networks \citep{wang2016duelingnetworkarchitecturesdeep} by simply introducing regularization in the function approximator like LayerNorm \citep{ba2016layernormalization} or BatchNorm \citep{ioffe2015batchnormalizationacceleratingdeep}. Coupled with this architectural change, PQN exploits vectorized environments by collecting experience in parallel for $T$ steps.

As our action space is continuous, we modify PQN through bang-off-bang control, treating continuous control as a multi-agent problem where each of the $M$ actuators is an agent in a cooperative game following \citet{seyde2023solvingcontinuouscontrolqlearning}. Then, the state-action function $Q_\theta(\bf s_t, a_t)$ is factorized as the average of M different state-action functions $Q_\theta^i(\mathbf{s_t},a_t^i)$, where the $i$th state-action function predicts the value of the bang-off-bang actions in $i$th action dimension following \citet{sunehag2017valuedecompositionnetworkscooperativemultiagent}. 
\begin{align}
    Q_\theta (\mathbf{s_t,a_t)}) =\frac{1}{M} \sum_{i=1}^M Q_\theta^i(\mathbf{s_t},a_t^i)
\end{align} 
In code for the Meta-World setting, the output of the state-action function is of size $(B,M,n_b)$ where $B$ is the batch size, $m$ is the action dimension/number of actuators (4) and $n_b$ is the number of bins per dimension (3). The action value is recovered by first taking the max over the bin dimension and then the mean over the action dimension. By taking the max over the bin dimension, \citet{seyde2023solvingcontinuouscontrolqlearning} sidestepped taking a max over the continuous action space. Now, we can compute the Bellman target and in the case of PQN, n-step returns.
\begin{align}
    y_t = r(\mathbf{s_t,a_t}) + \gamma \frac{1}{M} \sum_{i=1}^M \max_{a_{t+1}^i} Q_\theta^i(\mathbf{s_{t+1}},a_{t+1}^i)
\end{align}
\section{Meta-World}

\subsection{Success}
\label{appendix:meta-world-success}
We report two evaluation metrics, the overall success rate averaged across tasks and the cumulative reward achieved by the multi-task policy. Following the original Meta-World, success is a boolean indicating whether the robot brings the object within an $\epsilon$ distance of the goal position at \textit{any} point during the episode, which is less restrictive than works qualifying a success only if it occurs at the \textit{end} of an episode. Mathematically, success occurs if $\|o-g\|_2 < \epsilon$ is satisfied at least once, where $o$ is the object position and $g$ is the goal position.

Rather than defining the success rate as the maximum success rate over some evaluation rollouts as some previous work did, the success rate is defined as the proportion of success in the large number of environments that terminate their episodes every step. The reported success rate is this success rate averaged over the last 5 epochs of training. Due to massive parallelization, there is no need to separately roll out the learned policy in a separate process.
\subsection{More Results}
\label{appendix:addition_results}

\begin{table}[h!]
\centering
\resizebox{.95\columnwidth}{!}{
\begin{tabular}{lcccc}
\toprule
\multirow{2}{*}{\diagbox{Methods}{Tasks}} & \multicolumn{2}{c}{MT10-rand} & \multicolumn{2}{c}{MT50-rand} \\
\cmidrule(lr){2-3} \cmidrule(lr){4-5} 
 & SR $\uparrow$ & R $\uparrow$ & SR $\uparrow$ & R $\uparrow$ \\
\midrule
Vanilla & $87.51\,[86.99, 87.97]$ & $1032.99\,[1016.82, 1045.94]$ & $63.26\,[60.91, 65.37]$ & $817.77\, [789.13, 842.97]$ \\
Multihead & $85.19\,[81.42, 88.21]$ & $1005.69\,[980.87, 1027.55]$ & $74.03\,[71.47, 76.58]$ & $954.97\, [939.72, 962.84]$ \\
GRPO-Vanilla & $91.12\, [88.32, 93.96]$ & $916.32\, [899.83, 933.60]$ & $74.48\, [73.31, 75.64]$ & $916.83\, [898.69, 935.66]$ \\
\midrule
PCGrad & $86.21\,[83.19, 88.32]$ & $1038.27\, [1022.88, 1050.59]$ & $59.74\, [55.52, 64.12]$ & $760.99\, [739.05, 772.13]$ \\
CAGrad & $82.98\,[79.23, 86.27]$ & $938.43\, [896.83, 972.29]$ & $67.70\, [64.76, 70.53]$ & $874.45\, [845.62, 903.10]$ \\
FAMO & $87.26\,[82.53, 91.57]$ & $1016.11\, [964.30, 1053.88]$ & $74.52\, [73.25, 75.75]$ & $961.03\, [946.64, 976.15]$ \\
\midrule
PaCo & $84.37\,[81.61, 86.61]$ & $995.39\, [970.20, 1017.21]$ & $70.46\, [67.01, 73.32]$ & $917.84\, [881.61, 953.14]$ \\
SH-MOORE & $84.60\,[81.55, 87.59]$ & $1022.64\, [1006.23, 1037.54]$ & $66.33\, [64.56, 68.29]$ & $837.70\, [815.00, 860.89]$ \\
MH-MOORE & $86.94\,[83.91, 89.01]$ & $1044.85\, [1029.76, 1056.96]$ & $79.46\, [77.40, 82.24]$ & $1019.59\, [999.24, 1048.88]$ \\
SH-CARE & $81.51\,[78.52, 84.49]$ & $964.28\, [948.59, 979.89]$ & $67.51\, [66.33, 68.72]$ & $842.04 \,[822.31, 864.66]$ \\
MH-CARE & $84.79\,[81.34, 87.32]$ & $990.03\, [972.35, 1006.34]$ & $71.05\, [69.88, 72.30]$ & $863.88 \,[850.43, 878.51]$ \\
Soft-Modularization & $82.96\, [80.15, 85.66]$ & $994.29\, [980.24, 1009.03]$ & $67.72\, [65.06, 69.93]$ & $860.41\, [832.44, 883.77]$ \\
\bottomrule
\end{tabular}
}
\caption{95\% bootstrapped confidence intervals of the Meta-World evaluation metrics used to generate Figure \ref{fig: main_result_metaworld} and Figure \ref{fig:grpo_vs_vanilla} }
\label{table:main_table_results_metaworld}
\end{table}

\begin{figure}[h!]      
    \centering
    \includegraphics[width=.95\columnwidth]{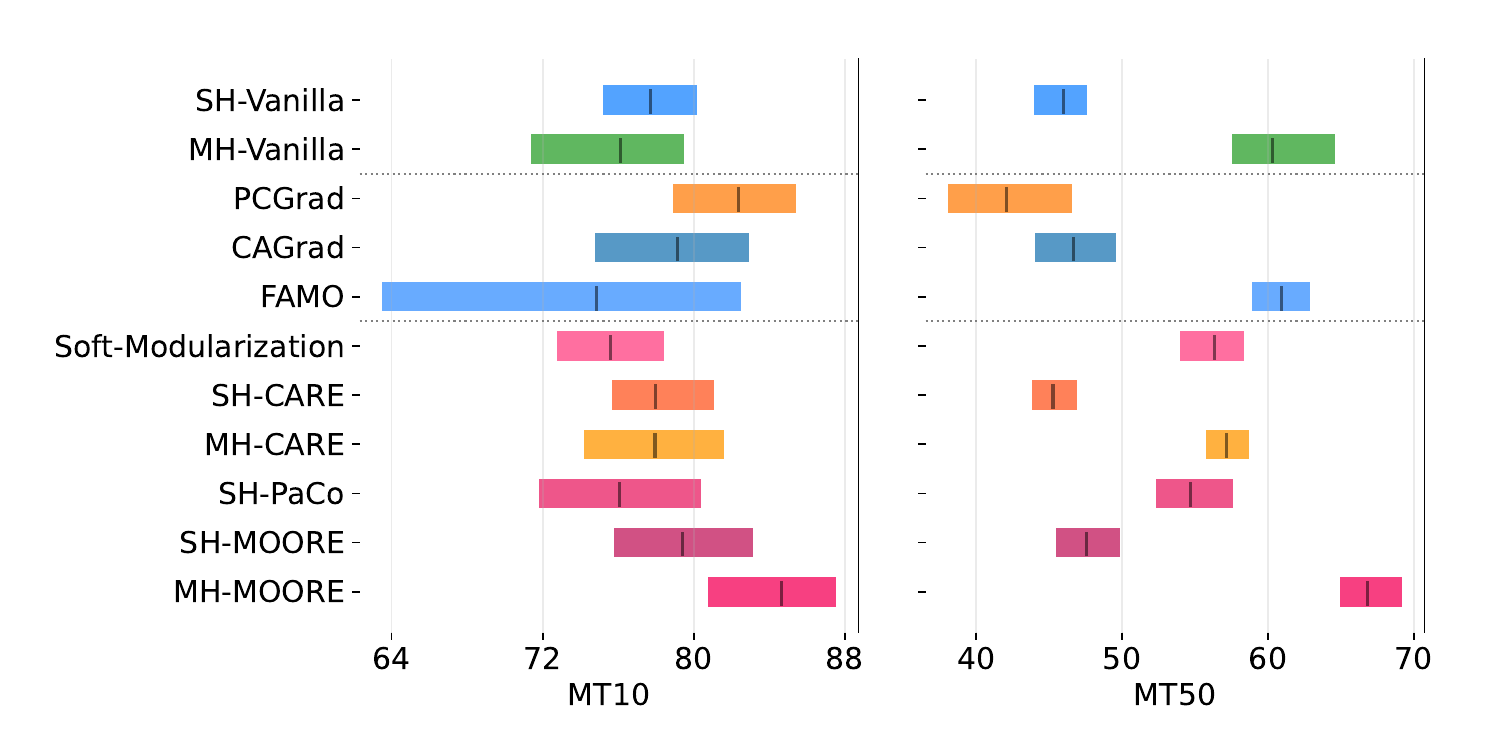}
    \caption{95\% bootstrapped CIs of the average success rate of all MT-PPO MTRL approaches using \textit{250M} frames per run over 10 seeds in Meta-World.  }
    \label{fig: main_result_low_data_regime}
\end{figure}

\subsubsection*{Acknowledgments}
\label{sec:ack}
This work has taken place in the Learning Agents Research
Group (LARG) at the Artificial Intelligence Laboratory, The University
of Texas at Austin.  LARG research is supported in part by the
National Science Foundation (FAIN-2019844, NRT-2125858), the Office of
Naval Research (N00014-18-2243), Army Research Office
(W911NF-23-2-0004, W911NF-17-2-0181), Lockheed Martin, and Good
Systems, a research grand challenge at the University of Texas at
Austin.  The views and conclusions contained in this document are
those of the authors alone.  Peter Stone serves as the Executive
Director of Sony AI America and receives financial compensation for
this work.  The terms of this arrangement have been reviewed and
approved by the University of Texas at Austin in accordance with its
policy on objectivity in research.


\bibliography{main}

\begin{thebibliography}{74}
\providecommand{\natexlab}[1]{#1}
\providecommand{\url}[1]{\texttt{#1}}
\expandafter\ifx\csname urlstyle\endcsname\relax
  \providecommand{\doi}[1]{DOI: #1}\else
  \providecommand{\doi}{DOI: \begingroup \urlstyle{rm}\Url}\fi

\bibitem[Agarwal et~al.(2021)Agarwal, Schwarzer, Castro, Courville, and Bellemare]{agarwal2021deep}
Rishabh Agarwal, Max Schwarzer, Pablo~Samuel Castro, Aaron~C Courville, and Marc Bellemare.
\newblock Deep reinforcement learning at the edge of the statistical precipice.
\newblock \emph{Advances in neural information processing systems}, 34:\penalty0 29304--29320, 2021.

\bibitem[Al-Hafez et~al.(2023)Al-Hafez, Zhao, Peters, and Tateo]{alhafez2023b}
Firas Al-Hafez, Guoping Zhao, Jan Peters, and Davide Tateo.
\newblock Locomujoco: A comprehensive imitation learning benchmark for locomotion.
\newblock In \emph{6th Robot Learning Workshop, NeurIPS}, 2023.

\bibitem[Allshire et~al.(2021)Allshire, Mittal, Lodaya, Makoviychuk, Makoviichuk, Widmaier, W{\"u}thrich, Bauer, Handa, and Garg]{allshire2021transferring}
Arthur Allshire, Mayank Mittal, Varun Lodaya, Viktor Makoviychuk, Denys Makoviichuk, Felix Widmaier, Manuel W{\"u}thrich, Stefan Bauer, Ankur Handa, and Animesh Garg.
\newblock Transferring dexterous manipulation from gpu simulation to a remote real-world trifinger.
\newblock \emph{arXiv preprint arXiv:2108.09779}, 2021.

\bibitem[Ba et~al.(2016)Ba, Kiros, and Hinton]{ba2016layernormalization}
Jimmy~Lei Ba, Jamie~Ryan Kiros, and Geoffrey~E. Hinton.
\newblock Layer normalization, 2016.
\newblock URL \url{https://arxiv.org/abs/1607.06450}.

\bibitem[Babaeizadeh et~al.(2016)Babaeizadeh, Frosio, Tyree, Clemons, and Kautz]{babaeizadeh2016reinforcement}
Mohammad Babaeizadeh, Iuri Frosio, Stephen Tyree, Jason Clemons, and Jan Kautz.
\newblock Reinforcement learning through asynchronous advantage actor-critic on a gpu.
\newblock \emph{arXiv preprint arXiv:1611.06256}, 2016.

\bibitem[Bonnet et~al.(2023)Bonnet, Luo, Byrne, Surana, Abramowitz, Duckworth, Coyette, Midgley, Tegegn, Kalloniatis, et~al.]{bonnet2023jumanji}
Cl{\'e}ment Bonnet, Daniel Luo, Donal Byrne, Shikha Surana, Sasha Abramowitz, Paul Duckworth, Vincent Coyette, Laurence~I Midgley, Elshadai Tegegn, Tristan Kalloniatis, et~al.
\newblock Jumanji: a diverse suite of scalable reinforcement learning environments in jax.
\newblock \emph{arXiv preprint arXiv:2306.09884}, 2023.

\bibitem[Chen et~al.(2022)Chen, Xu, Yu, Li, Ma, Xu, and Hsu]{chen2022daxbench}
Siwei Chen, Yiqing Xu, Cunjun Yu, Linfeng Li, Xiao Ma, Zhongwen Xu, and David Hsu.
\newblock Daxbench: Benchmarking deformable object manipulation with differentiable physics.
\newblock \emph{arXiv preprint arXiv:2210.13066}, 2022.

\bibitem[Cheng et~al.(2024)Cheng, Shi, Agarwal, and Pathak]{cheng2024extreme}
Xuxin Cheng, Kexin Shi, Ananye Agarwal, and Deepak Pathak.
\newblock Extreme parkour with legged robots.
\newblock In \emph{2024 IEEE International Conference on Robotics and Automation (ICRA)}, pp.\  11443--11450. IEEE, 2024.

\bibitem[Cobbe et~al.(2020)Cobbe, Hesse, Hilton, and Schulman]{cobbe2020leveraging}
Karl Cobbe, Chris Hesse, Jacob Hilton, and John Schulman.
\newblock Leveraging procedural generation to benchmark reinforcement learning.
\newblock In \emph{International conference on machine learning}, pp.\  2048--2056. PMLR, 2020.

\bibitem[D'Oro et~al.(2022)D'Oro, Schwarzer, Nikishin, Bacon, Bellemare, and Courville]{d2022sample}
Pierluca D'Oro, Max Schwarzer, Evgenii Nikishin, Pierre-Luc Bacon, Marc~G Bellemare, and Aaron Courville.
\newblock Sample-efficient reinforcement learning by breaking the replay ratio barrier.
\newblock In \emph{Deep Reinforcement Learning Workshop NeurIPS 2022}, 2022.

\bibitem[Espeholt et~al.(2018)Espeholt, Soyer, Munos, Simonyan, Mnih, Ward, Doron, Firoiu, Harley, Dunning, et~al.]{espeholt2018impala}
Lasse Espeholt, Hubert Soyer, Remi Munos, Karen Simonyan, Vlad Mnih, Tom Ward, Yotam Doron, Vlad Firoiu, Tim Harley, Iain Dunning, et~al.
\newblock Impala: Scalable distributed deep-rl with importance weighted actor-learner architectures.
\newblock In \emph{International conference on machine learning}, pp.\  1407--1416. PMLR, 2018.

\bibitem[{Franka Robotics}(2017)]{FrankaPanda}
{Franka Robotics}.
\newblock Franka emika panda robot, 2017.
\newblock URL \url{https://www.franka.de}.
\newblock Accessed: 2025-02-17.

\bibitem[Freeman et~al.(2021)Freeman, Frey, Raichuk, Girgin, Mordatch, and Bachem]{freeman2021brax}
C~Daniel Freeman, Erik Frey, Anton Raichuk, Sertan Girgin, Igor Mordatch, and Olivier Bachem.
\newblock Brax--a differentiable physics engine for large scale rigid body simulation.
\newblock \emph{arXiv preprint arXiv:2106.13281}, 2021.

\bibitem[Fu et~al.(2023)Fu, Cheng, and Pathak]{fu2023deep}
Zipeng Fu, Xuxin Cheng, and Deepak Pathak.
\newblock Deep whole-body control: learning a unified policy for manipulation and locomotion.
\newblock In \emph{Conference on Robot Learning}, pp.\  138--149. PMLR, 2023.

\bibitem[Gallici et~al.(2024)Gallici, Fellows, Ellis, Pou, Masmitja, Foerster, and Martin]{gallici2024simplifyingdeeptemporaldifference}
Matteo Gallici, Mattie Fellows, Benjamin Ellis, Bartomeu Pou, Ivan Masmitja, Jakob~Nicolaus Foerster, and Mario Martin.
\newblock Simplifying deep temporal difference learning, 2024.
\newblock URL \url{https://arxiv.org/abs/2407.04811}.

\bibitem[Haarnoja et~al.(2018)Haarnoja, Zhou, Abbeel, and Levine]{haarnoja2018softactorcriticoffpolicymaximum}
Tuomas Haarnoja, Aurick Zhou, Pieter Abbeel, and Sergey Levine.
\newblock Soft actor-critic: Off-policy maximum entropy deep reinforcement learning with a stochastic actor, 2018.
\newblock URL \url{https://arxiv.org/abs/1801.01290}.

\bibitem[Hendawy et~al.(2024)Hendawy, Peters, and D'Eramo]{hendawy2024multitaskreinforcementlearningmixture}
Ahmed Hendawy, Jan Peters, and Carlo D'Eramo.
\newblock Multi-task reinforcement learning with mixture of orthogonal experts, 2024.
\newblock URL \url{https://arxiv.org/abs/2311.11385}.

\bibitem[Hessel et~al.(2019)Hessel, Soyer, Espeholt, Czarnecki, Schmitt, and Van~Hasselt]{hessel2019multi}
Matteo Hessel, Hubert Soyer, Lasse Espeholt, Wojciech Czarnecki, Simon Schmitt, and Hado Van~Hasselt.
\newblock Multi-task deep reinforcement learning with popart.
\newblock In \emph{Proceedings of the AAAI Conference on Artificial Intelligence}, volume~33, pp.\  3796--3803, 2019.

\bibitem[Horgan et~al.(2018)Horgan, Quan, Budden, Barth-Maron, Hessel, van Hasselt, and Silver]{horgan2018distributedprioritizedexperiencereplay}
Dan Horgan, John Quan, David Budden, Gabriel Barth-Maron, Matteo Hessel, Hado van Hasselt, and David Silver.
\newblock Distributed prioritized experience replay, 2018.
\newblock URL \url{https://arxiv.org/abs/1803.00933}.

\bibitem[Hwangbo et~al.(2019)Hwangbo, Lee, Dosovitskiy, Bellicoso, Tsounis, Koltun, and Hutter]{hwangbo2019learning}
Jemin Hwangbo, Joonho Lee, Alexey Dosovitskiy, Dario Bellicoso, Vassilios Tsounis, Vladlen Koltun, and Marco Hutter.
\newblock Learning agile and dynamic motor skills for legged robots.
\newblock \emph{Science Robotics}, 4\penalty0 (26):\penalty0 eaau5872, 2019.

\bibitem[Ioffe \& Szegedy(2015)Ioffe and Szegedy]{ioffe2015batchnormalizationacceleratingdeep}
Sergey Ioffe and Christian Szegedy.
\newblock Batch normalization: Accelerating deep network training by reducing internal covariate shift, 2015.
\newblock URL \url{https://arxiv.org/abs/1502.03167}.

\bibitem[James et~al.(2020)James, Ma, Rovick~Arrojo, and Davison]{james2019rlbench}
Stephen James, Zicong Ma, David Rovick~Arrojo, and Andrew~J. Davison.
\newblock Rlbench: The robot learning benchmark \& learning environment.
\newblock \emph{IEEE Robotics and Automation Letters}, 2020.

\bibitem[Kirk et~al.(2023)Kirk, Zhang, Grefenstette, and Rockt{\"a}schel]{kirk2023survey}
Robert Kirk, Amy Zhang, Edward Grefenstette, and Tim Rockt{\"a}schel.
\newblock A survey of zero-shot generalisation in deep reinforcement learning.
\newblock \emph{Journal of Artificial Intelligence Research}, 76:\penalty0 201--264, 2023.

\bibitem[Koyamada et~al.(2023)Koyamada, Okano, Nishimori, Murata, Habara, Kita, and Ishii]{koyamada2023pgx}
Sotetsu Koyamada, Shinri Okano, Soichiro Nishimori, Yu~Murata, Keigo Habara, Haruka Kita, and Shin Ishii.
\newblock Pgx: Hardware-accelerated parallel game simulators for reinforcement learning.
\newblock \emph{Advances in Neural Information Processing Systems}, 36:\penalty0 45716--45743, 2023.

\bibitem[Lange(2022)]{gymnax2022github}
Robert~Tjarko Lange.
\newblock {gymnax}: A {JAX}-based reinforcement learning environment library, 2022.
\newblock URL \url{http://github.com/RobertTLange/gymnax}.

\bibitem[Lee et~al.(2020)Lee, Hwangbo, Wellhausen, Koltun, and Hutter]{lee2020learning}
Joonho Lee, Jemin Hwangbo, Lorenz Wellhausen, Vladlen Koltun, and Marco Hutter.
\newblock Learning quadrupedal locomotion over challenging terrain.
\newblock \emph{Science robotics}, 5\penalty0 (47):\penalty0 eabc5986, 2020.

\bibitem[Li et~al.(2023)Li, Chen, Hong, Ajay, and Agrawal]{li2023parallel}
Zechu Li, Tao Chen, Zhang-Wei Hong, Anurag Ajay, and Pulkit Agrawal.
\newblock Parallel $q$-learning: Scaling off-policy reinforcement learning under massively parallel simulation.
\newblock In \emph{International Conference on Machine Learning}. PMLR, 2023.

\bibitem[Liang et~al.(2018{\natexlab{a}})Liang, Liaw, Nishihara, Moritz, Fox, Goldberg, Gonzalez, Jordan, and Stoica]{liang2018rllib}
Eric Liang, Richard Liaw, Robert Nishihara, Philipp Moritz, Roy Fox, Ken Goldberg, Joseph Gonzalez, Michael Jordan, and Ion Stoica.
\newblock Rllib: Abstractions for distributed reinforcement learning.
\newblock In \emph{International conference on machine learning}, pp.\  3053--3062. PMLR, 2018{\natexlab{a}}.

\bibitem[Liang et~al.(2018{\natexlab{b}})Liang, Makoviychuk, Handa, Chentanez, Macklin, and Fox]{liang2018gpu}
Jacky Liang, Viktor Makoviychuk, Ankur Handa, Nuttapong Chentanez, Miles Macklin, and Dieter Fox.
\newblock Gpu-accelerated robotic simulation for distributed reinforcement learning.
\newblock In \emph{Conference on Robot Learning}, pp.\  270--282. PMLR, 2018{\natexlab{b}}.

\bibitem[Liang et~al.(2024)Liang, Wang, Wang, Bastani, Jayaraman, and Ma]{liang2024eurekaverseenvironmentcurriculumgeneration}
William Liang, Sam Wang, Hung-Ju Wang, Osbert Bastani, Dinesh Jayaraman, and Yecheng~Jason Ma.
\newblock Eurekaverse: Environment curriculum generation via large language models, 2024.
\newblock URL \url{https://arxiv.org/abs/2411.01775}.

\bibitem[Liu et~al.(2023{\natexlab{a}})Liu, Feng, Stone, and Liu]{liu2023famofastadaptivemultitask}
Bo~Liu, Yihao Feng, Peter Stone, and Qiang Liu.
\newblock Famo: Fast adaptive multitask optimization, 2023{\natexlab{a}}.
\newblock URL \url{https://arxiv.org/abs/2306.03792}.

\bibitem[Liu et~al.(2023{\natexlab{b}})Liu, Zhu, Gao, Feng, Liu, Zhu, and Stone]{liu2023liberobenchmarkingknowledgetransfer}
Bo~Liu, Yifeng Zhu, Chongkai Gao, Yihao Feng, Qiang Liu, Yuke Zhu, and Peter Stone.
\newblock Libero: Benchmarking knowledge transfer for lifelong robot learning, 2023{\natexlab{b}}.
\newblock URL \url{https://arxiv.org/abs/2306.03310}.

\bibitem[Liu et~al.(2024)Liu, Liu, Jin, Stone, and Liu]{liu2024conflictaversegradientdescentmultitask}
Bo~Liu, Xingchao Liu, Xiaojie Jin, Peter Stone, and Qiang Liu.
\newblock Conflict-averse gradient descent for multi-task learning, 2024.
\newblock URL \url{https://arxiv.org/abs/2110.14048}.

\bibitem[Luo et~al.(2023)Luo, Cao, Kitani, Xu, et~al.]{luo2023perpetual}
Zhengyi Luo, Jinkun Cao, Kris Kitani, Weipeng Xu, et~al.
\newblock Perpetual humanoid control for real-time simulated avatars.
\newblock In \emph{Proceedings of the IEEE/CVF International Conference on Computer Vision}, pp.\  10895--10904, 2023.

\bibitem[Makoviichuk \& Makoviychuk(2021)Makoviichuk and Makoviychuk]{rl-games2021}
Denys Makoviichuk and Viktor Makoviychuk.
\newblock rl-games: A high-performance framework for reinforcement learning.
\newblock \url{https://github.com/Denys88/rl_games}, May 2021.

\bibitem[Makoviychuk et~al.(2021)Makoviychuk, Wawrzyniak, Guo, Lu, Storey, Macklin, Hoeller, Rudin, Allshire, Handa, and State]{makoviychuk2021isaacgymhighperformance}
Viktor Makoviychuk, Lukasz Wawrzyniak, Yunrong Guo, Michelle Lu, Kier Storey, Miles Macklin, David Hoeller, Nikita Rudin, Arthur Allshire, Ankur Handa, and Gavriel State.
\newblock Isaac gym: High performance gpu-based physics simulation for robot learning, 2021.
\newblock URL \url{https://arxiv.org/abs/2108.10470}.

\bibitem[Margolis et~al.(2024)Margolis, Yang, Paigwar, Chen, and Agrawal]{margolis2024rapid}
Gabriel~B Margolis, Ge~Yang, Kartik Paigwar, Tao Chen, and Pulkit Agrawal.
\newblock Rapid locomotion via reinforcement learning.
\newblock \emph{The International Journal of Robotics Research}, 43\penalty0 (4):\penalty0 572--587, 2024.

\bibitem[Matthews et~al.(2024)Matthews, Beukman, Ellis, Samvelyan, Jackson, Coward, and Foerster]{matthews2024craftaxlightningfastbenchmarkopenended}
Michael Matthews, Michael Beukman, Benjamin Ellis, Mikayel Samvelyan, Matthew Jackson, Samuel Coward, and Jakob Foerster.
\newblock Craftax: A lightning-fast benchmark for open-ended reinforcement learning, 2024.
\newblock URL \url{https://arxiv.org/abs/2402.16801}.

\bibitem[Mittal et~al.(2023)Mittal, Yu, Yu, Liu, Rudin, Hoeller, Yuan, Singh, Guo, Mazhar, Mandlekar, Babich, State, Hutter, and Garg]{mittal2023orbit}
Mayank Mittal, Calvin Yu, Qinxi Yu, Jingzhou Liu, Nikita Rudin, David Hoeller, Jia~Lin Yuan, Ritvik Singh, Yunrong Guo, Hammad Mazhar, Ajay Mandlekar, Buck Babich, Gavriel State, Marco Hutter, and Animesh Garg.
\newblock Orbit: A unified simulation framework for interactive robot learning environments.
\newblock \emph{IEEE Robotics and Automation Letters}, 8\penalty0 (6):\penalty0 3740--3747, 2023.
\newblock \doi{10.1109/LRA.2023.3270034}.

\bibitem[Mnih et~al.(2013)Mnih, Kavukcuoglu, Silver, Graves, Antonoglou, Wierstra, and Riedmiller]{mnih2013playingatarideepreinforcement}
Volodymyr Mnih, Koray Kavukcuoglu, David Silver, Alex Graves, Ioannis Antonoglou, Daan Wierstra, and Martin Riedmiller.
\newblock Playing atari with deep reinforcement learning, 2013.
\newblock URL \url{https://arxiv.org/abs/1312.5602}.

\bibitem[Mnih et~al.(2015)Mnih, Kavukcuoglu, Silver, Rusu, Veness, Bellemare, Graves, Riedmiller, Fidjeland, Ostrovski, Petersen, Beattie, Sadik, Antonoglou, King, Kumaran, Wierstra, Legg, and Hassabis]{Mnih2015HumanlevelCT}
Volodymyr Mnih, Koray Kavukcuoglu, David Silver, Andrei~A. Rusu, Joel Veness, Marc~G. Bellemare, Alex Graves, Martin~A. Riedmiller, Andreas~Kirkeby Fidjeland, Georg Ostrovski, Stig Petersen, Charlie Beattie, Amir Sadik, Ioannis Antonoglou, Helen King, Dharshan Kumaran, Daan Wierstra, Shane Legg, and Demis Hassabis.
\newblock Human-level control through deep reinforcement learning.
\newblock \emph{Nature}, 518:\penalty0 529--533, 2015.
\newblock URL \url{https://api.semanticscholar.org/CorpusID:205242740}.

\bibitem[Mnih et~al.(2016)Mnih, Badia, Mirza, Graves, Lillicrap, Harley, Silver, and Kavukcuoglu]{mnih2016asynchronous}
Volodymyr Mnih, Adria~Puigdomenech Badia, Mehdi Mirza, Alex Graves, Timothy Lillicrap, Tim Harley, David Silver, and Koray Kavukcuoglu.
\newblock Asynchronous methods for deep reinforcement learning.
\newblock In \emph{International conference on machine learning}, pp.\  1928--1937. PmLR, 2016.

\bibitem[Morad et~al.(2023)Morad, Kortvelesy, Bettini, Liwicki, and Prorok]{morad2023popgym}
Steven Morad, Ryan Kortvelesy, Matteo Bettini, Stephan Liwicki, and Amanda Prorok.
\newblock Popgym: Benchmarking partially observable reinforcement learning.
\newblock \emph{arXiv preprint arXiv:2303.01859}, 2023.

\bibitem[Nair et~al.(2015)Nair, Srinivasan, Blackwell, Alcicek, Fearon, De~Maria, Panneershelvam, Suleyman, Beattie, Petersen, et~al.]{nair2015massively}
Arun Nair, Praveen Srinivasan, Sam Blackwell, Cagdas Alcicek, Rory Fearon, Alessandro De~Maria, Vedavyas Panneershelvam, Mustafa Suleyman, Charles Beattie, Stig Petersen, et~al.
\newblock Massively parallel methods for deep reinforcement learning.
\newblock \emph{arXiv preprint arXiv:1507.04296}, 2015.

\bibitem[OpenAI et~al.(2019)OpenAI, Andrychowicz, Baker, Chociej, Jozefowicz, McGrew, Pachocki, Petron, Plappert, Powell, Ray, Schneider, Sidor, Tobin, Welinder, Weng, and Zaremba]{openai2019learningdexterousinhandmanipulation}
OpenAI, Marcin Andrychowicz, Bowen Baker, Maciek Chociej, Rafal Jozefowicz, Bob McGrew, Jakub Pachocki, Arthur Petron, Matthias Plappert, Glenn Powell, Alex Ray, Jonas Schneider, Szymon Sidor, Josh Tobin, Peter Welinder, Lilian Weng, and Wojciech Zaremba.
\newblock Learning dexterous in-hand manipulation, 2019.
\newblock URL \url{https://arxiv.org/abs/1808.00177}.

\bibitem[Park et~al.(2024)Park, Frans, Eysenbach, and Levine]{park2024ogbench}
Seohong Park, Kevin Frans, Benjamin Eysenbach, and Sergey Levine.
\newblock Ogbench: Benchmarking offline goal-conditioned rl.
\newblock \emph{arXiv preprint arXiv:2410.20092}, 2024.

\bibitem[Petrenko et~al.(2020)Petrenko, Huang, Kumar, Sukhatme, and Koltun]{petrenko2020sample}
Aleksei Petrenko, Zhehui Huang, Tushar Kumar, Gaurav Sukhatme, and Vladlen Koltun.
\newblock Sample factory: Egocentric 3d control from pixels at 100000 fps with asynchronous reinforcement learning.
\newblock In \emph{International Conference on Machine Learning}, pp.\  7652--7662. PMLR, 2020.

\bibitem[Pinto \& Gupta(2016)Pinto and Gupta]{pinto2016learningpushgraspingusing}
Lerrel Pinto and Abhinav Gupta.
\newblock Learning to push by grasping: Using multiple tasks for effective learning, 2016.
\newblock URL \url{https://arxiv.org/abs/1609.09025}.

\bibitem[Rudin et~al.(2022)Rudin, Hoeller, Reist, and Hutter]{rudin2022learning}
Nikita Rudin, David Hoeller, Philipp Reist, and Marco Hutter.
\newblock Learning to walk in minutes using massively parallel deep reinforcement learning.
\newblock In \emph{Conference on Robot Learning}, pp.\  91--100. PMLR, 2022.

\bibitem[Rutherford et~al.(2024)Rutherford, Ellis, Gallici, Cook, Lupu, Ingvarsson, Willi, Hammond, Khan, de~Witt, Souly, Bandyopadhyay, Samvelyan, Jiang, Lange, Whiteson, Lacerda, Hawes, Rocktaschel, Lu, and Foerster]{rutherford2024jaxmarlmultiagentrlenvironments}
Alexander Rutherford, Benjamin Ellis, Matteo Gallici, Jonathan Cook, Andrei Lupu, Gardar Ingvarsson, Timon Willi, Ravi Hammond, Akbir Khan, Christian~Schroeder de~Witt, Alexandra Souly, Saptarashmi Bandyopadhyay, Mikayel Samvelyan, Minqi Jiang, Robert~Tjarko Lange, Shimon Whiteson, Bruno Lacerda, Nick Hawes, Tim Rocktaschel, Chris Lu, and Jakob~Nicolaus Foerster.
\newblock Jaxmarl: Multi-agent rl environments and algorithms in jax, 2024.
\newblock URL \url{https://arxiv.org/abs/2311.10090}.

\bibitem[Schulman et~al.(2017)Schulman, Wolski, Dhariwal, Radford, and Klimov]{schulman2017proximalpolicyoptimizationalgorithms}
John Schulman, Filip Wolski, Prafulla Dhariwal, Alec Radford, and Oleg Klimov.
\newblock Proximal policy optimization algorithms, 2017.
\newblock URL \url{https://arxiv.org/abs/1707.06347}.

\bibitem[Seyde et~al.(2023)Seyde, Werner, Schwarting, Gilitschenski, Riedmiller, Rus, and Wulfmeier]{seyde2023solvingcontinuouscontrolqlearning}
Tim Seyde, Peter Werner, Wilko Schwarting, Igor Gilitschenski, Martin Riedmiller, Daniela Rus, and Markus Wulfmeier.
\newblock Solving continuous control via q-learning, 2023.
\newblock URL \url{https://arxiv.org/abs/2210.12566}.

\bibitem[Sferrazza et~al.(2024)Sferrazza, Huang, Lin, Lee, and Abbeel]{sferrazza2024humanoidbench}
Carmelo Sferrazza, Dun-Ming Huang, Xingyu Lin, Youngwoon Lee, and Pieter Abbeel.
\newblock Humanoidbench: Simulated humanoid benchmark for whole-body locomotion and manipulation.
\newblock \emph{arXiv preprint arXiv:2403.10506}, 2024.

\bibitem[Shao et~al.(2024)Shao, Wang, Zhu, Xu, Song, Bi, Zhang, Zhang, Li, Wu, and Guo]{shao2024deepseekmathpushinglimitsmathematical}
Zhihong Shao, Peiyi Wang, Qihao Zhu, Runxin Xu, Junxiao Song, Xiao Bi, Haowei Zhang, Mingchuan Zhang, Y.~K. Li, Y.~Wu, and Daya Guo.
\newblock Deepseekmath: Pushing the limits of mathematical reasoning in open language models, 2024.
\newblock URL \url{https://arxiv.org/abs/2402.03300}.

\bibitem[Silver et~al.(2016)Silver, Huang, Maddison, Guez, Sifre, Van Den~Driessche, Schrittwieser, Antonoglou, Panneershelvam, Lanctot, et~al.]{silver2016mastering}
David Silver, Aja Huang, Chris~J Maddison, Arthur Guez, Laurent Sifre, George Van Den~Driessche, Julian Schrittwieser, Ioannis Antonoglou, Veda Panneershelvam, Marc Lanctot, et~al.
\newblock Mastering the game of go with deep neural networks and tree search.
\newblock \emph{nature}, 529\penalty0 (7587):\penalty0 484--489, 2016.

\bibitem[Singla et~al.(2024)Singla, Agarwal, and Pathak]{sapg2024}
Jayesh Singla, Ananye Agarwal, and Deepak Pathak.
\newblock Sapg: Split and aggregate policy gradients.
\newblock In \emph{Proceedings of the 41st International Conference on Machine Learning (ICML 2024)}, Proceedings of Machine Learning Research, Vienna, Austria, July 2024. PMLR.

\bibitem[Sodhani \& Zhang(2021)Sodhani and Zhang]{Sodhani2021MTRL}
Shagun Sodhani and Amy Zhang.
\newblock Mtrl - multi task rl algorithms.
\newblock Github, 2021.
\newblock URL \url{https://github.com/facebookresearch/mtrl}.

\bibitem[Sodhani et~al.(2021)Sodhani, Zhang, and Pineau]{Sodhani2021MultiTaskRL}
Shagun Sodhani, Amy Zhang, and Joelle Pineau.
\newblock Multi-task reinforcement learning with context-based representations.
\newblock In \emph{International Conference on Machine Learning}, 2021.
\newblock URL \url{https://api.semanticscholar.org/CorpusID:231879645}.

\bibitem[Sun et~al.(2022)Sun, Zhang, Xu, and Tomizuka]{Sun2022PaCoPM}
Lingfeng Sun, Haichao Zhang, Wei Xu, and Masayoshi Tomizuka.
\newblock Paco: Parameter-compositional multi-task reinforcement learning.
\newblock \emph{ArXiv}, abs/2210.11653, 2022.
\newblock URL \url{https://api.semanticscholar.org/CorpusID:253080666}.

\bibitem[Sunehag et~al.(2017)Sunehag, Lever, Gruslys, Czarnecki, Zambaldi, Jaderberg, Lanctot, Sonnerat, Leibo, Tuyls, and Graepel]{sunehag2017valuedecompositionnetworkscooperativemultiagent}
Peter Sunehag, Guy Lever, Audrunas Gruslys, Wojciech~Marian Czarnecki, Vinicius Zambaldi, Max Jaderberg, Marc Lanctot, Nicolas Sonnerat, Joel~Z. Leibo, Karl Tuyls, and Thore Graepel.
\newblock Value-decomposition networks for cooperative multi-agent learning, 2017.
\newblock URL \url{https://arxiv.org/abs/1706.05296}.

\bibitem[Sutton et~al.(1999)Sutton, McAllester, Singh, and Mansour]{sutton1999policy}
Richard~S Sutton, David McAllester, Satinder Singh, and Yishay Mansour.
\newblock Policy gradient methods for reinforcement learning with function approximation.
\newblock \emph{Advances in neural information processing systems}, 12, 1999.

\bibitem[Tao et~al.(2024)Tao, Xiang, Shukla, Qin, Hinrichsen, Yuan, Bao, Lin, Liu, kai Chan, Gao, Li, Mu, Xiao, Gurha, Huang, Calandra, Chen, Luo, and Su]{taomaniskill3}
Stone Tao, Fanbo Xiang, Arth Shukla, Yuzhe Qin, Xander Hinrichsen, Xiaodi Yuan, Chen Bao, Xinsong Lin, Yulin Liu, Tse kai Chan, Yuan Gao, Xuanlin Li, Tongzhou Mu, Nan Xiao, Arnav Gurha, Zhiao Huang, Roberto Calandra, Rui Chen, Shan Luo, and Hao Su.
\newblock Maniskill3: Gpu parallelized robotics simulation and rendering for generalizable embodied ai.
\newblock \emph{arXiv preprint arXiv:2410.00425}, 2024.

\bibitem[Tassa et~al.(2018)Tassa, Doron, Muldal, Erez, Li, de~Las~Casas, Budden, Abdolmaleki, Merel, Lefrancq, Lillicrap, and Riedmiller]{tassa2018deepmindcontrolsuite}
Yuval Tassa, Yotam Doron, Alistair Muldal, Tom Erez, Yazhe Li, Diego de~Las~Casas, David Budden, Abbas Abdolmaleki, Josh Merel, Andrew Lefrancq, Timothy Lillicrap, and Martin Riedmiller.
\newblock Deepmind control suite, 2018.
\newblock URL \url{https://arxiv.org/abs/1801.00690}.

\bibitem[Todorov et~al.(2012)Todorov, Erez, and Tassa]{Mujoco}
Emanuel Todorov, Tom Erez, and Yuval Tassa.
\newblock Mujoco: A physics engine for model-based control.
\newblock In \emph{2012 IEEE/RSJ International Conference on Intelligent Robots and Systems}, pp.\  5026--5033, 2012.
\newblock \doi{10.1109/IROS.2012.6386109}.

\bibitem[{Unitree Robotics}(2021)]{unitree_go1}
{Unitree Robotics}.
\newblock \emph{Go1 User Manual}.
\newblock Unitree Robotics, 2021.
\newblock Available at https://www.unitree.com/go1.

\bibitem[Wang et~al.(2016)Wang, Schaul, Hessel, van Hasselt, Lanctot, and de~Freitas]{wang2016duelingnetworkarchitecturesdeep}
Ziyu Wang, Tom Schaul, Matteo Hessel, Hado van Hasselt, Marc Lanctot, and Nando de~Freitas.
\newblock Dueling network architectures for deep reinforcement learning, 2016.
\newblock URL \url{https://arxiv.org/abs/1511.06581}.

\bibitem[Wurman et~al.(2022)Wurman, Barrett, Kawamoto, MacGlashan, Subramanian, Walsh, Capobianco, Devlic, Eckert, Fuchs, et~al.]{wurman2022outracing}
Peter~R Wurman, Samuel Barrett, Kenta Kawamoto, James MacGlashan, Kaushik Subramanian, Thomas~J Walsh, Roberto Capobianco, Alisa Devlic, Franziska Eckert, Florian Fuchs, et~al.
\newblock Outracing champion gran turismo drivers with deep reinforcement learning.
\newblock \emph{Nature}, 602\penalty0 (7896):\penalty0 223--228, 2022.

\bibitem[Xing et~al.(2024)Xing, Luk, and Oh]{xing2024stabilizing}
Eliot Xing, Vernon Luk, and Jean Oh.
\newblock Stabilizing reinforcement learning in differentiable multiphysics simulation.
\newblock \emph{arXiv preprint arXiv:2412.12089}, 2024.

\bibitem[Yang et~al.(2020)Yang, Xu, Wu, and Wang]{yang2020multitaskreinforcementlearningsoft}
Ruihan Yang, Huazhe Xu, Yi~Wu, and Xiaolong Wang.
\newblock Multi-task reinforcement learning with soft modularization, 2020.
\newblock URL \url{https://arxiv.org/abs/2003.13661}.

\bibitem[Yu et~al.(2020)Yu, Kumar, Gupta, Levine, Hausman, and Finn]{yu2020gradientsurgerymultitasklearning}
Tianhe Yu, Saurabh Kumar, Abhishek Gupta, Sergey Levine, Karol Hausman, and Chelsea Finn.
\newblock Gradient surgery for multi-task learning, 2020.
\newblock URL \url{https://arxiv.org/abs/2001.06782}.

\bibitem[Yu et~al.(2021)Yu, Quillen, He, Julian, Narayan, Shively, Bellathur, Hausman, Finn, and Levine]{yu2021metaworldbenchmarkevaluationmultitask}
Tianhe Yu, Deirdre Quillen, Zhanpeng He, Ryan Julian, Avnish Narayan, Hayden Shively, Adithya Bellathur, Karol Hausman, Chelsea Finn, and Sergey Levine.
\newblock Meta-world: A benchmark and evaluation for multi-task and meta reinforcement learning, 2021.
\newblock URL \url{https://arxiv.org/abs/1910.10897}.

\bibitem[Zakka et~al.(2025)Zakka, Tabanpour, Liao, Haiderbhai, Holt, Luo, Allshire, Frey, Sreenath, Kahrs, Sferrazza, Tassa, and Abbeel]{mujoco_playground_2025}
Kevin Zakka, Baruch Tabanpour, Qiayuan Liao, Mustafa Haiderbhai, Samuel Holt, Jing~Yuan Luo, Arthur Allshire, Erik Frey, Koushil Sreenath, Lueder~A. Kahrs, Carlo Sferrazza, Yuval Tassa, and Pieter Abbeel.
\newblock Mujoco playground: An open-source framework for gpu-accelerated robot learning and sim-to-real transfer., 2025.
\newblock URL \url{https://github.com/google-deepmind/mujoco_playground}.

\bibitem[Zhu et~al.(2020)Zhu, Wong, Mandlekar, Mart{\'\i}n-Mart{\'\i}n, Joshi, Nasiriany, and Zhu]{zhu2020robosuite}
Yuke Zhu, Josiah Wong, Ajay Mandlekar, Roberto Mart{\'\i}n-Mart{\'\i}n, Abhishek Joshi, Soroush Nasiriany, and Yifeng Zhu.
\newblock robosuite: A modular simulation framework and benchmark for robot learning.
\newblock \emph{arXiv preprint arXiv:2009.12293}, 2020.

\bibitem[Zhuang et~al.(2023)Zhuang, Fu, Wang, Atkeson, Schwertfeger, Finn, and Zhao]{zhuang2023robot}
Ziwen Zhuang, Zipeng Fu, Jianren Wang, Christopher Atkeson, Soeren Schwertfeger, Chelsea Finn, and Hang Zhao.
\newblock Robot parkour learning.
\newblock \emph{arXiv preprint arXiv:2309.05665}, 2023.

\end{thebibliography}
\bibliographystyle{rlj}

\beginSupplementaryMaterials
\section{MTRL Approaches}
\label{appendix:mtrl_overview}
Here, we present an overview of each state-of-the-art MTRL baseline in MTBench. 

\subsection{Gradient manipulation methods}
Gradient manipulation methods compute a new gradient of the multi-task objective, incurring the overhead of solving an optimization problem per iteration as well as storing and computing $K$ task gradients.

\paragraph{PCGrad:} Projecting Conflicting Gradients \citep{yu2020gradientsurgerymultitasklearning} observe when the gradients of any two task objectives $l_i$ conflict ( defined as having negative cosine similarity) and when their magnitudes are sufficiently different, optimization using the average gradient will cause negative transfer. It attempts to resolve gradient confliction by a simple procedure manipulating each task gradient $\nabla l_i$ to be the result of iteratively removing the conflict with each task gradient $\nabla l_j$, $\forall j \in [K], j\neq i$.
\begin{align}
    \nabla l_i' \leftarrow \nabla l_i - \frac{\nabla l_i^T \nabla l_j}{\|\nabla l_j\|^2} \nabla l_j & \quad \text{if} \quad \nabla l_i^T \nabla l_j < 0
\end{align}

\paragraph{CAGrad: } Conflict-Averse Gradient descent \citep{liu2024conflictaversegradientdescentmultitask} resolves the gradient conflict by finding an update vector $d \in \mathbb{R}^m$ that minimizes the worst-case gradient conflict across all the tasks. More specifically, let $g_i$ be the gradient of task $i \in [K]$, and $g_0$ be the gradient computed from the average loss, CAGrad seeks to solve such an optimization problem:
\begin{align}
\max_{d \in \mathbb{R}^m} \min_{i \in [K]} \langle g_i, d \rangle \quad \text{s.t.} \quad \|d - g_0\| \leq c \|g_0\|
\end{align}
Here, $c \in [0, 1)$ is a pre-specified hyper-parameter that controls the convergence rate. The optimization problem looks for the best update vector within a local ball centered at the averaged gradient $g_0$, which also minimizes the conflict in losses $\langle g_i, d \rangle$.

\paragraph{FAMO: } Fast Adaptive Multitask Optimization \citep{liu2023famofastadaptivemultitask} addresses the under-optimization of certain tasks when using standard gradient descent on averaged losses without incurring the $O(K)$ cost to compute and store all task gradients, which can be significant, especially as the number of tasks increases. FAMO leverages loss history to adaptively adjust task weights, ensuring balanced optimization across tasks while maintaining $O(1)$ space and time complexity per iteration.

\subsection{Neural Architectures} 
Neural Architecture methods seek to avoid task interference by learning shared representations, which are fed to the prediction head. Such representations accelerate MTRL.

\paragraph{CARE: } Contextual Attention-based Representation learning \citep{Sodhani2021MTRL} utilizes metadata associated with the set of tasks to weight the representations learned by a mixture of encoders through the attention mechanism.
\paragraph{MOORE: } Mixture Of Orthogonal Experts \citep{hendawy2024multitaskreinforcementlearningmixture} uses a mixture of experts to encode the state and orthogonalizes those representations to encourage diversity, weighting these representations from a task encoder. 
\paragraph{PaCo: } Parameter
Compositional \citep{Sun2022PaCoPM} learns a base parameter set $\phi = [\phi_1 \cdots \phi_k]$ and task-specific compositional vector $w_k$ such that multiplying $\phi$ and $w_k$ represents the task  parameters $\theta_k$. 
\paragraph{Soft-Modularization: } \citet{yang2020multitaskreinforcementlearningsoft} also uses a mixture of experts to encode the state but also uses a routing network to softly combine the outputs at each layer based on the task.

\section{Hyperparameter Details}
In this section, we provide hyperparameter values for each MTRL approach.
\begin{table}[bth!]
\resizebox{1.0\columnwidth}{!}{%
\begin{tabular}{l@{\hspace{30pt}}l@{\hspace{30pt}}l}
\toprule
Description & value & variable\_name \\
\midrule
Number of environments & 24576 / 24576 & num\_envs \\
Network hidden sizes & [256,128,64] & network.mlp.units \\
Minibatch size & 16384 / 32768 & minibatch\_size \\
Horizon length & 32 & horizon \\
Mini-epochs & 5 & mini\_epochs \\
Number of epochs & 1272 / 1272 & max\_epochs \\
Episode length & 150 & episodeLength \\
Discount factor & 0.99 & gamma \\
Clip ratio & 0.2 & e\_clip \\
Policy entropy coefficient & .005 & entropy\_coef \\
Optimizer learning rate & 5e-4 & learning\_rate \\
Optimizer learning schedule & fixed & lr\_schedule \\
Advantage estimation tau & 0.95 & tau\\
Value Normalization by task & True & normalize\_value\\
Input Normalization by task & True & normalize\_input\\
Separate critic and policy networks & True & network.separate \\
\midrule
CARE-Specific Hyperparameters \\
\midrule
Network hidden sizes & [400,400,400] & care.units \\
Mixture of Encoders experts & 6 & encoder.num\_experts \\
Mixture of Encoders layers & 2 & encoder.num\_layers \\
Mixture of Encoders hidden dim & 50 & encoder.D \\
Attention temperature & 1.0 & encoder.temperature \\
Post-Attention MLP hidden sizes & [50,50] & attention.units \\
Context encoder hidden sizes & [50,50] & context\_encoder.units \\
Context encoder bias & True & context\_encoder.bias  \\
\midrule
MOORE-Specific Hyperparameters \\
\midrule
MoE experts & 4 / 6  & moore.num\_experts \\
MoE layers & 3 & moore.num\_layers \\
MoE hidden dim & 400 & moore.D \\
Activation before/after task encoding weighting & [Linear, Tanh] & moore.agg\_activation \\
Task encoder hidden sizes & [256] & task\_encoder.units \\
Task encoder bias & False & task\_encoder.bias \\
\midrule
PaCo-Specific Hyperparameters\\
\midrule
Number of Compositional Vectors & 5 / 20 & paco.K \\
Network hidden dim & 400 & paco.D \\
Network layers & 3 & paco.num\_layers \\
Task encoder bias & False & task\_encoder.bias \\
Task encoder init & orthogonal & task\_encoder.compositional\_initializer \\
Task encoder activation & softmax & task\_encoder.activation \\

\midrule
Soft-Modularization-Specific  Hyperparameters \\
\midrule
MoE experts & 2 & soft\_network.num\_experts \\
MoE layers & 4 & soft\_network.num\_layer \\
State encoder hidden sizes & [256,256] & state\_encoder.units \\
Task encoder hidden sizes & [256] & task\_encoder.units\\

\midrule
PCGrad  Hyperparameters \\
\midrule
Number of environments & 24576 / 8192 & num\_envs \\
Project actor gradient & False & project\_actor\_gradient \\
Project critic gradient & True & project\_critic\_gradient \\

\midrule
CAGrad  Hyperparameters \\
\midrule
Number of environments & 24576 / 6144 & num\_envs \\
Project actor gradient & False & project\_actor\_gradient \\
Project critic gradient & True & project\_critic\_gradient \\
Local ball radius for searching update vector & 0.4 & c \\
\midrule
FAMO  Hyperparameters \\
\midrule
Regularization coefficient & 1e-3 & gamma \\
Learning rate of the task logits & 1e-3 & w\_lr \\
Clipping value of the task logits & 1e-2 & epsilon \\
Normalize the task logits gradients & True & norm\_w\_grad \\
\bottomrule
\end{tabular}
}
\caption{Hyperparameters used for MTPPO. A '/' indicates the value used for Meta-World's MT10/MT50 respectively, and otherwise is identical for each setting. }
\label{table:hyperparameters}
\end{table}

\begin{table}[bth!]
\resizebox{1.0\columnwidth}{!}{%
\begin{tabular}{l@{\hspace{30pt}}l@{\hspace{30pt}}l}
\toprule
Description & value & variable\_name \\
\midrule
Number of environments & 4096 / 24576 & num\_envs \\
Minibatch size & 16384 / 76800 & minibatch\_size \\
Episode length & 150 & episodeLength \\
Horizon length & 150 & horizon \\
Mini-epochs & 5 & mini\_epochs \\
Number of epochs & 1908 / 1272 & max\_epochs \\
Discount factor & 0.99 & gamma \\
Clip ratio & 0.2 & e\_clip \\
Policy entropy coefficient & .005 & entropy\_coef \\
Optimizer learning rate & 5e-4 & learning\_rate \\
Optimizer learning schedule & fixed & lr\_schedule \\
Advantage estimation tau & 0.95 & tau\\
Value Normalization by task & True & normalize\_value\\
Input Normalization by task & True & normalize\_input\\
Separate critic and policy networks & True & network.separate \\
\bottomrule
\end{tabular}
}
\caption{Hyperparameters used for MT-GRPO in MT10 / MT50. A '/' indicates the value used for MT10/MT50 respectively and otherwise is identical for each setting.}
\label{table:hyperparameters}
\end{table}

\begin{table}[htb!]
\resizebox{1.0\columnwidth}{!}{
\begin{tabular}{l@{\hspace{30pt}}l@{\hspace{30pt}}l}
\toprule
Description & value & variable\_name \\
\midrule
Number of environments & 8192 & num\_envs \\
Gamma & .99 & gamma \\
Peng's Q(lambda) & .5 & q\_lambda \\
Number of minibatches & 4 & num\_minibatches \\
Episode length & 500 & episodeLength \\
Bang-off-Bang & 3 & binsPerDim \\
Action Scale & .005 & actionScale \\
Mini epochs & 8 & mini\_epochs \\
Max grad norm & 10.0 & max\_grad\_norm \\
Horizon & 16 & horizon \\
Start epsilon & 1.0 & start\_e \\
End epsilon & 0.005 & end\_e \\
Decay epsilon & True & decay\_epsilon \\
Fraction of exploration steps & .005 & exploration\_fraction \\
Critic learning rate & 3e-4 & critic\_lr \\
Anneal learning rate & True & anneal\_lr \\
Value Normalization by task & False & normalize\_value\\
Input Normalization by task & False & normalize\_input\\
Use residual connections & True & q.residual\_network \\
Number of LayerNormAndResidualMLPs & 2 & q.num\_blocks \\
Network hidden dim & 256 & q.D \\
Batch norm input & False & q.norm\_first\_layer\\
\bottomrule
\end{tabular}
}
\caption{Hyperparameters used for MT-PQN in MT10.}
\label{table:hyperparameters}
\end{table}

\begin{table}[htb!]
\resizebox{1.0\columnwidth}{!}{%
\begin{tabular}{l@{\hspace{30pt}}l@{\hspace{30pt}}l}
\toprule
Description & value & variable\_name \\
\midrule
Number of environments & 4096 & num\_envs \\
Network hidden sizes & [512,256,128] & network.mlp.units \\
Gamma & .99 & gamma \\
Separate critic and policy networks & True & network.separate \\
Number of Gradient steps per epoch & 32 & gradient\_steps\_per\_itr \\
Learnable temperature & True & learnable\_ temperature \\ 
Use distangeled alpha & True & use\_disentangled\_alpha \\
Initial alpha & 1 & init\_alpha \\
Alpha learning rate & 5e-3 & alpha\_lr \\
Critic learning rate & 5e-4 & critic\_lr \\
Critic tau & .01 & critic\_tau \\
Batch size & 8192 & batch\_size \\
N-step reward & 16 & nstep \\
Grad norm & .5 & grad\_norm \\
Horizon & 1 & horizon \\
Value Normalization by task & True & normalize\_value\\
Input Normalization by task & True & normalize\_input\\
Replay Buffer Size & 5000000 & replay\_buffer\_size \\
Target entropy coef & 1.0 & target\_entropy\_coef\\
\bottomrule
\end{tabular}
}
\caption{Hyperparameters used for MT-SAC in MT10/MT50. A '/' indicates the value used for MT10/MT50 respectively and otherwise is identical for each setting. MT-SAC is very sensitive to the number of environments and replay ratio in the massively parallel regime.}
\label{table:hyperparameters}
\end{table}

\begin{table}[htb!]
\resizebox{1.0\columnwidth}{!}{%
\begin{tabular}{l@{\hspace{30pt}}l@{\hspace{30pt}}l}
\toprule
Description & value & variable\_name \\
\midrule
Minibatch size & 16384 & minibatch\_size \\
Horizon length & 32 & horizon \\
Mini-epochs & 5 & mini\_epochs \\
Number of epochs & 2000 / 4000 & max\_epochs \\
Episode length & 800 &  \\
Discount factor & 0.99 & gamma \\
Clip ratio & 0.2 & e\_clip \\
Policy entropy coefficient & .005 & entropy\_coef \\
Optimizer learning rate & 5e-4 & learning\_rate \\
Optimizer learning schedule & adaptive & lr\_schedule \\
Advantage estimation tau & 0.95 & tau\\
Value Normalization by task & False & normalize\_value\\
Input Normalization by task & False & normalize\_input\\
Separate critic and policy networks & True & network.separate \\
\midrule
MOORE-Specific Hyperparameters \\
\midrule
MoE experts & 2  & moore.num\_experts \\
MoE layers & 2 & moore.num\_layers \\
MoE hidden dim & 256 & moore.D \\
Activation before/after task encoding weighting & [Linear, Linear] & moore.agg\_activation \\
Task encoder hidden sizes & [128] & \\
Task encoder bias & False & task\_encoder.bias \\
Multihead & False & multi\_head \\
\midrule
PaCo-Specific Hyperparameters\\
\midrule
Number of Compositional Vectors & 5 & paco.K \\
Network hidden dim & 400 & paco.D \\
Network layers & 3 & paco.num\_layers \\
Task encoder bias & False & task\_encoder.bias \\
Task encoder init & orthogonal & task\_encoder.compositional\_initializer \\
Task encoder activation & softmax & task\_encoder.activation \\
\midrule
Soft-Modularization-Specific  Hyperparameters \\
\midrule
MoE experts & 2 & soft\_network.num\_experts \\
MoE layers & 2 & soft\_network.num\_layer \\
State encoder hidden sizes & [256,256] & state\_encoder.units \\
Task encoder hidden sizes & [128] & task\_encoder.units\\
\midrule
PCGrad  Hyperparameters \\
\midrule
Project actor gradient & False & project\_actor\_gradient \\
Project critic gradient & True & project\_critic\_gradient \\
\midrule
CAGrad  Hyperparameters \\
\midrule
Project actor gradient & False & project\_actor\_gradient \\
Project critic gradient & True & project\_critic\_gradient \\
Local ball radius for searching update vector & 0.4 & c \\
\midrule
FAMO  Hyperparameters \\
\midrule
Regularization coefficient & 1e-4 & gamma \\
Learning rate of the task logits & 5e-3 & w\_lr \\
Small value for the clipping of the task logits & 1e-3 & epsilon \\
Normalize the task logits gradients & True & norm\_w\_grad \\
\bottomrule
\end{tabular}
}
\caption{Hyperparameters used for MT-PPO in Parkour Benchmark. A '/' indicates the value used for Parkour-easy/Parkour-hard respectively and otherwise is identical for each setting.}
\label{table:hyperparameters}
\end{table}







\end{document}